\documentclass[letterpaper,journal]{IEEEtran}
\usepackage{amsmath,amsfonts}
\usepackage{algorithmic}
\usepackage{algorithm}
\usepackage{array}
\usepackage[caption=false,font=normalsize,labelfont=sf,textfont=sf]{subfig}
\usepackage{textcomp}
\usepackage{stfloats}
\usepackage{url}
\usepackage[hidelinks]{hyperref}
\usepackage{verbatim}
\usepackage{graphicx}
\usepackage{cite}
\usepackage{xcolor}  
\usepackage{booktabs} 
\usepackage{tabularx} 
\usepackage{multirow}
\usepackage{colortbl} 
\usepackage{makecell} 
\usepackage{orcidlink}
\usepackage{xcolor}

\begin{document}
	
	\title{MicroSDF: Microfacet-Driven Hybrid Neural SDFs for Mixed-Reflectance Surface Reconstruction}
	
	\author{Lejia Ye$^{\orcidlink{0009-0007-5189-2589}}$,
		Yuhua Xu, 
		Yulan Guo$^{\orcidlink{0000-0001-7051-841X}}$,~\IEEEmembership{Senior Member, IEEE}, 
		Lian Xu$^{\orcidlink{0000-0002-1759-2941}}$, 
		Xiaohu Zhang$^{\orcidlink{0000-0003-4907-1451}}$
		\thanks{Published in IEEE Transactions on Image Processing, vol.~35, pp.~4106--4121, Apr.~2026, doi: \url{https://doi.org/10.1109/TIP.2026.3678096}.}%
		\thanks{
			Lejia Ye and Xiaohu Zhang are with the School of Aeronautics and Astronautics, Sun Yat-sen University, Shenzhen Campus, Shenzhen
			518107, China. (e-mail: yelj8@mail2.sysu.edu.cn;zhangxiaohu@mail.sysu.edu.cn.)
			
			Yuhua Xu is with Orbbec Inc., Orbbec Technology Building, No. 88 High-tech North 1st Road, Nanshan District, Shenzhen 518052, China(e-mail: xuyuhua@orbbec.com).
			
			Yulan Guo is with the School of Electronics and Communication Engineering, Sun Yat-sen University, Shenzhen Campus, Shenzhen
			518107, China (e-mail: guoyulan@sysu.edu.cn)
			
			Lian Xu is with the Department of Computer Science and Software Engineering, The University of Western Australia, Perth, WA 6009, Australia (e-mail: lian.xu@uwa.edu.au).
			
			\textit{(Corresponding author: Xiaohu Zhang and Lian Xu.)}
		}%
	}
	
	\maketitle
	
	\begin{abstract}
		Accurate 3D reconstruction in real-world environments remains a significant challenge due to the coexistence of reflective and non-reflective surfaces, which pose distinct modeling demands. Existing methods often treat these surface types separately, limiting their generalizability and physical plausibility.
		To bridge this gap, we propose MicroSDF, a novel neural implicit framework that facilitates geometry and reflectance modeling through microfacet theory. Our approach incorporates three core innovations: (\textbf{i}) a microfacet-guided geometry model that extracts multi-scale surface normals (macroscopic and microfacet) from a signed distance field (SDF), regularized by a proposed microfacet normal consistency loss to enforce physically plausible surface orientations; (\textbf{ii}) an enhanced dual-branch color model, where the specular branch leverages the microfacet normals to model high-frequency reflectance, and the vanilla branch, unlike prior works, uses reflection direction (instead of viewing direction) to better model diffuse and low-frequency specular components;
		and (\textbf{iii}) a detection-guided color blending strategy that adaptively fuses the color outputs based on reflection priors, providing more physically intuitive blending than implicitly learned blending weights. Combined with a tailored multi-stage optimization scheme, the proposed MicroSDF achieves robust and high-fidelity reconstruction across reflective and non-reflective surfaces. Extensive experiments on DTU, Shiny Blender, Ref-NeRF, and DeepVoxels datasets demonstrate state-of-the-art performance, establishing a new direction for physically grounded neural reconstruction.
	\end{abstract}

	
	\section{Introduction}
	\IEEEPARstart{3}D surface reconstruction aims to recover the geometry of a scene from 2D images, enabling downstream tasks such as novel view synthesis and scene understanding. 
	It is a fundamental task in computer vision~\cite{hartley2003multiple}, with widespread applications in robotics, autonomous navigation, and augmented reality. Real-world scenes often contain a mixture of reflective and non-reflective surfaces, coexisting within the same environment.
	However, most existing methods are designed to handle either diffuse (non-reflective)~\cite{barnes2009patchmatch,NEURIPS2021_e41e164f}, or highly reflective surfaces~\cite{ge2023ref-neus}, independently. This limits their generalization in mixed scenes, reducing their robustness in practical applications. To achieve high-fidelity and reliable reconstruction in diverse real-world scenarios, a unified framework that adaptively models both surface types is essential.

	\begin{figure}[t]
		\centering
		\includegraphics[width=1.0\linewidth]{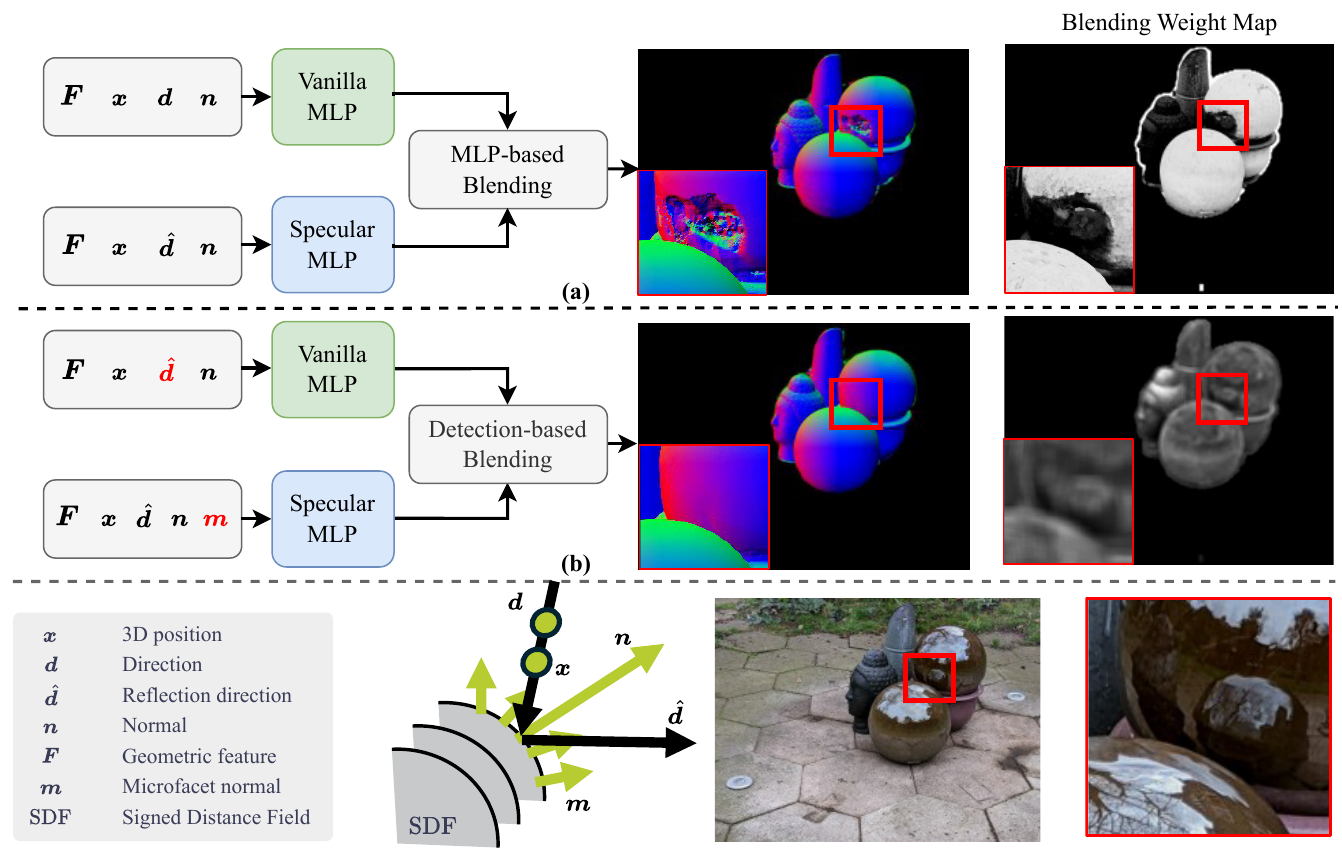}
		\vspace{-0.5cm}
		\caption{Comparison between prior works and ours for implicit 3D reconstruction.
			(\textbf{a}) Prior dual-branch color modeling methods~\cite{wang2023unisdf,gao2024anisdf}. (\textbf{b}) The proposed dual-branch color modeling method.  Our approach differs from prior works in the following three aspects: (\textbf{i}) Using reflection direction $\hat{\boldsymbol{d}}$ as input to the vanilla branch to better model diffuse and low-frequency specular reflections, (\textbf{ii}) leveraging microfacet normals as input to the specular branch to capture high-frequency specular highlights, and (\textbf{iii}) using detection-based weights for physically intuitive color blending. 
			These designs lead to significantly better geometric reconstruction on complex reflective surfaces than prior works.
		}
		\vspace{-0.5cm}
		\label{fig:the-differences}
	\end{figure}
	Current reconstruction paradigms face significant limitations when applied to reflective surfaces.
	For example, traditional Multi-View Stereo (MVS) methods rely on the assumption of multi-view consistency, which is commonly violated in reflective regions due to their strong view-dependent appearance.  
	While recent neural implicit representations, especially neural Signed Distance Field (SDF) methods~\cite{NEURIPS2021_e41e164f, li2023neuralangelo,wang2023neus2}, have significantly advanced 3D reconstruction, their underlying rendering formulations often neglect the physical properties of highly reflective surfaces. Consequently, these methods struggle to disentangle view-dependent appearance from true geometry in reflective regions, leading to inaccurate reconstructions. 
	
	The difficulty of handling reflective surfaces is also encountered in inverse rendering, which aims to infer the scene's geometry, material properties, and lighting from observed images. This is, however, fundamentally ill-posed: multiple combinations of material parameters, lighting, and shape can produce the same observed color, making decomposition from color supervision solely inherently ambiguous~\cite{guo2014robust, ge2023ref-neus,zhang2021physg}.
	To address this, recent inverse rendering approaches~\cite{boss2021nerd,srinivasan2021nerv,zhang2021nerfactor} have incorporated physically grounded reflectance models, enabling more plausible and interpretable rendering results. 
	While our goal differs, these advances raise a natural question: \textit{can physical principles be similarly integrated into neural implicit reconstruction frameworks to better handle reflective surfaces?} 
	Inspired by this, we turn to microfacet theory~\cite{pharr2023physically}, a foundational model in graphics that captures surface reflectance behavior through statistical distributions of microscopic surface normals. A key insight from this theory is the distinction between reflective and non-reflective regions: specular regions exhibit small variance in microfacet normals, while non-specular regions show large variance.
	
	Building upon this insight, we propose MicroSDF, a unified neural SDF framework that explicitly embeds microfacet-inspired priors into the reconstruction process, for handling both reflective and non reflective surfaces. It inherits the coarse-to-fine learning paradigm of recent neural SDF methods \cite{wang2023neus2,hugoycj_instant_angelo,li2023neuralangelo,NEURIPS2021_e41e164f,yariv2021volume,yariv2023bakedsdf,fu2022geo,ge2023ref-neus} , with sequential geometry and color networks. More specifically, MicroSDF introduces three key innovations: 
	(\textbf{i}) \textbf{Microfacet-guided geometry modeling}. 
	We sample local surface normals from a finer-scale geometry hierarchy around each query point and concatenate them with the numerical normal at the current level. To ensure physically consistent reflectance behavior, we introduce a microfacet normal consistency loss that encourages alignment between the microfacet normals and local surface normal in reflective regions, enabling coherent reflection directions.
	(\textbf{ii}) \textbf{Microfacet-driven color modeling}. We employ a dual-branch color network, as illustrated in Fig.~\ref{fig:the-differences}, in contrast to prior methods~\cite{wang2023unisdf,gao2024anisdf}, our specular branch incorporates microfacet normals as additional inputs to model high-frequency reflective behavior, and the vanilla branch uses the reflection direction as input to better capture diffuse and low-frequency reflective effects. This design ensures that these two branches focus on complementary aspects of color modeling. 
	(\textbf{iii}) \textbf{Detection-guided color blending}. While the specular branch enables reflectance modeling, correctly identifying where to apply it remains challenging. Prior works~\cite{wang2023unisdf} have used implicitly learned blending weights from an auxiliary MLP, which often fail to reliably identify specular highlights from true geometry.
	In contrast, we propose a detection-guided blending module that leverages per-pixel specular likelihoods from a pretrained detector (\textit{e.g.}, StableDelight~\cite{stableDelight2025}) to guide the fusion of color predictions from the two branches. This enables physically intuitive color blending, reducing artifacts and improving reconstruction robustness across mixed-material scenes.
	
	Through these designs, the proposed MicroSDF achieves robust and high-fidelity reconstruction of reflective and non-reflective surfaces within a unified framework. The proposed method consistently improves performance in challenging reflective regions while preserving accuracy on standard scenes. In summary, our main contributions are fourfold: 
	\begin{enumerate}
		\item We propose \textbf{MicroSDF}, the first neural implicit surface reconstruction framework that explicitly integrates microfacet theory in a principled manner.
		The proposed MicroSDF offers a unified solution to robustly handle complex real-world scenes with both reflective and non-reflective surfaces.
		
		\item We propose a microfacet normal consistency loss that aligns microfacet and local surface normals in specular regions, enhancing physically meaningful geometry that supports accurate view-dependent appearance.
		
		\item We propose an enhanced dual-branch color model that incorporates microfacet normals into the specular branch to capture high-frequency reflections, and uses reflection direction in the vanilla branch to model diffuse and low-frequency reflections. Guided by a detection-based reflectance prior, this design enables robust and physically intuitive color blending.
		
		\item The proposed MicroSDF achieves the state-of-the-art performance across diverse benchmarks, including DTU, Shiny Blender, Ref-NeRF, and DeepVoxels, showing the effectiveness on both reflective and non-reflective surfaces.
	\end{enumerate}
	
	\section{Related Works}
	\label{Related-Works}
	\subsection{Multi-view Stereo}
	Multi-view stereo (MVS) aims to reconstruct fine-grained scene geometry from multi-view images. 
	Traditional MVS methods can be classified into four categories based on the output scene representation: volumetric-based methods, mesh-based methods, point cloud-based methods, and depth map-based methods~\cite{seitz2006comparison}. 
	Among them, depth map-based methods\cite{schoenberger2016sfm, schoenberger2016mvs,shen2013accurate,li2020confidence} are the most flexible, estimating depth maps for each view by utilizing photometric consistency between reference and neighboring images, then fusing all the depth maps into dense point clouds. 
	A typical example is COLMAP~\cite{schoenberger2016sfm, schoenberger2016mvs}, which employs a robust patch-based matching strategy to estimate depth maps and then fuses them into a complete point cloud representation. Surface reconstruction methods, such as screened Poisson Surface Reconstruction~\cite{kazhdan2013screened}, are then employed to reconstruct surfaces from the point clouds.
	
	The advent of deep learning has transformed MVS by leveraging data-driven techniques for feature extraction and depth estimation. 
	MVSNet~\cite{yao2018mvsnet} pioneered deep learning in MVS by using CNNs to construct 3D cost volumes and regress depth maps. 
	Inspired by MVSNet, a series of outstanding works have emerged~\cite{chen2020mvsnet++,zhang2023multi,liu2023epipolar,10120723,chang2024ei}. 
	MVSNet++~\cite{chen2020mvsnet++} improved upon MVSNet by introducing feature pyramid structures and instance normalization, fusing multi-level information for accurate 3D point localization and generating smooth depth maps through tailored loss functions and a curriculum learning framework. 
	Subsequent works like RA-MVSNet~\cite{zhang2023multi} and ET-MVSNet~\cite{liu2023epipolar} enhanced accuracy through recurrent aggregation and epipolar constraints, while GeoMVSNet~\cite{zhang2023geomvsnet} incorporated geometric priors for better depth prediction. 
	NR-MVSNet~\cite{10120723} further refined the coarse-to-fine strategy with a normal consistency-based depth hypotheses module and a depth refinement module with reliable attention, producing smoother and more accurate depth maps, particularly in texture-less and repetitive-texture regions. 
	More recently, Transformer-based methods, such as MVSFormer~\cite{caomvsformer}, MVSFormer++~\cite{cao2024mvsformer++}, and EI-MVSNet~\cite{chang2024ei}, have improved performance by modeling long-range view dependencies, with EI-MVSNet leveraging epipolar-guided cost volume construction and an interval-aware depth estimation module to enhance depth estimation accuracy. 
	However, these deep learning-based MVS methods typically require large training datasets.
	
	While the above traditional and learning-based MVS methods have shown promising results in general scenarios, they face fundamental limitations when dealing with reflective surfaces. The core photometric consistency assumption underlying these methods breaks down for glossy materials, making accurate reconstruction challenging.
	
	\subsection{Neural Implicit Surface Based Reconstruction}
	
	In recent years, representation methods such as Neural Radiance Fields (NeRF) ~\cite{mildenhall2020nerf} and 3D Gaussian Splatting (3DGS)~\cite{kerbl20233d} have revolutionized traditional 3D reconstruction, advancing tasks such as novel view rendering~\cite{bao2025loopsparsegs,9725268}, human performance capture~\cite{11104970}, human pose and shape estimation~\cite{9709705}, object segmentation~\cite{zheng2024surface} and Multi-View 3D detection~\cite{huang2025nerf}. 
	
	3D Gaussian Splatting (3DGS)~\cite{kerbl20233d} offers high rendering quality and efficiency and has been widely adopted for novel view synthesis. However, achieving accurate geometry remains challenging because it lacks an explicit surface representation, which makes reliable surface extraction and geometry-dependent downstream tasks difficult. To improve geometric fidelity, several recent methods align Gaussian primitives with the underlying surface to enable surface extraction, including 2DGS~\cite{huang20242d}, BBSplat~\cite{svitov2025billboard}, SuGaR~\cite{guedon2024sugar}, GOF~\cite{yu2024gaussian} and Rade-GS~\cite{zhang2024rade}. However, their geometric accuracy in practice remains generally inferior to that of neural SDF based methods.
	Several other variants of 3D Gaussian splatting~\cite{li2025watersplatting,dai2025rainygs,zhou2025lita} have integrated domain-specific physical principles into the 3DGS framework for different objectives. For example, WaterSplatting~\cite{li2025watersplatting} combines 3DGS with an underwater light transport model, explicitly accounting for absorption and scattering to enable fast underwater scene reconstruction. RainyGS~\cite{dai2025rainygs} incorporates physically based raindrop and shallow-water simulations to synthesize dynamic rain effects such as water accumulation and surface waves in reconstructed scenes. LITA-GS~\cite{zhou2025lita} leverages Kubelka–Munk theory to introduce illumination-invariant physical priors, thereby facilitating illumination-agnostic novel view synthesis under challenging lighting conditions.
	In contrast to these 3DGS based methods that are tailored to specific tasks such as underwater reconstruction, rain synthesis and illumination-robust rendering, our work focuses on surface reconstruction and geometric accuracy by adopting an implicit SDF representation and employing microfacet theory as a geometric regularizer, with the explicit goal of achieving high-fidelity surfaces and accurate geometry.
	
	Native NeRF similarly faces geometric limitations: its reconstruction capabilities are constrained by inherent geometric-color ambiguity~\cite{zhang2020nerf++,fu2022geo} and the limitations of volumetric density representations, resulting in imprecise surface details. To address these limitations, subsequent research~\cite{fu2022geo,li2023neuralangelo,yariv2021volume,long2022sparseneus,oechsle2021unisurf,wang2023neus2,NEURIPS2021_e41e164f,yariv2023bakedsdf} has explored explicit surface representations (e.g., SDF-based formulations) to achieve high-precision geometry in NeRF-based methods.
	Among them, VolSDF~\cite{yariv2021volume} introduced a significant advancement by incorporating SDF with a Laplace transform, establishing a mathematical relationship between SDF and volumetric density. 
	This enables precise mesh extraction through Marching Cubes~\cite{Lorensen_2020}. 
	Building upon this, NeuralAngelo~\cite{li2023neuralangelo} has further refined the reconstruction pipeline, achieving enhanced surface fidelity and overall reconstruction quality.
	
	Numerous studies~\cite{deng2022depth, roessle2022dense, wei2021nerfingmvs,fu2022geo,zhang2024pg,yu2022monosdf} have demonstrated the potential for mutual enhancement through the integration of NeRF with traditional MVS approaches. 
	Among these, several methods~\cite{fu2022geo,zhang2024pg, yu2022monosdf} have incorporated external geometric priors, such as depth maps, point clouds, and normal maps, to provide intermediate supervision for SDF-based networks, thereby improving geometric reconstruction quality. 
	Notably, ~\cite{fu2022geo} identified a fundamental bias between rendered colors and geometry, and theoretically established that direct supervision of the zero-level set of SDF networks using point cloud information offers unbiased advantages and facilitates neural implicit surface learning. 
	However, the effectiveness of these approaches heavily relies on the quality of external supervision. 
	Both externally supervised geometric information and native NeRF techniques face challenges in reconstructing reflective surfaces, as NeRF's view-dependent color prediction mechanism, while effective for diffuse materials, fails to accurately model the complex light interactions and view-dependent effects of reflective surfaces. Furthermore, external supervision often relies on multi-view consistency assumptions, which may not hold in real-world scenarios.
	
	\subsection{Reflective Surface Reconstruction}
	
	Recent research in inverse rendering ~\cite{liang2023envidr, guo2022nerfren, boss2021neural, zhang2021nerfactor, jin2023tensoir,mai2023neural} has increasingly embraced physics-based rendering models to address the challenge of reflectance interpretation. 
	While these methods employ various basis functions to approximate the rendering equation, they primarily focus on improving rendering quality, without explicit investigation and evaluation of the geometry. 
	Notably, Neural Microfacet Fields (NMF)~\cite{mai2023neural} shares a similar high-level motivation with our work: both explicitly leverage microfacet-based physical modeling as priors to reduce the ambiguity of learning from images, particularly for reflective surfaces.
	However, NMF and our MicroSDF are designed for different primary tasks: NMF targets volumetric inverse rendering with explicit material–illumination decomposition, whereas MicroSDF focuses on high-precision geometric reconstruction. In terms of how microfacets are incorporated, NMF embeds microfacet modeling into a volumetric representation by assigning local normals and microfacet BRDF parameters to every volumetric sample along the ray, and performs surface-style Monte Carlo reflection integration to jointly recover materials and lighting and enable relighting, with geometry represented implicitly by a density field rather than an explicit SDF surface. In contrast, MicroSDF introduces microfacets only on the explicit SDF surface, defining microfacet normals at the surface that are used both as inputs to a specular branch to capture reflective appearance and as a consistency prior that regularizes surface orientation, thereby prioritizing accurate geometry recovery in scenes with mixed reflective and non-reflective regions without performing explicit material–illumination decomposition. 
	
	Ref-NeRF~\cite{2022refnerf} achieved a notable breakthrough by introducing an innovative reparametrization technique that effectively disentangles reflectance from appearance. Nevertheless, its geometric reconstructions remain susceptible to the complex nature of view-dependent effects. 
	Building upon the reparameterization strategy of Ref-NeRF, Ref-NeuS~\cite{ge2023ref-neus} integrated an anomaly detection module to effectively identify and handle specular regions.
	While this solution demonstrates effectiveness for reflective surfaces, it often underperforms when dealing with ordinary objects.
	
	UniSDF~\cite{wang2023unisdf} proposed a more universal solution by introducing a dual-branch architecture to handle both reflective and non-reflective regions. 
	However, the core idea of UniSDF remains constrained by complex lighting phenomena in real-world scenarios, such as near-field reflection, which often leads to geometric reconstructions with significant artifacts. Through three modifications illustrated in Fig.~\ref{fig:the-differences}, we significantly enhance its performance in such settings.
	More recently, AniSDF~\cite{gao2024anisdf} attempted to develop a universal reconstruction method by leveraging a dual-granularity geometric network and physics-based encoding. In contrast, our approach maintains a single geometric network by sampling normals at different scales. Furthermore, our modifications adopt a “loosely coupled” design with existing neural implicit reconstruction paradigms, facilitating easier practical application.
	
	
	
	\begin{figure*}[t]
		\centering
		\includegraphics[width=1.0\linewidth]{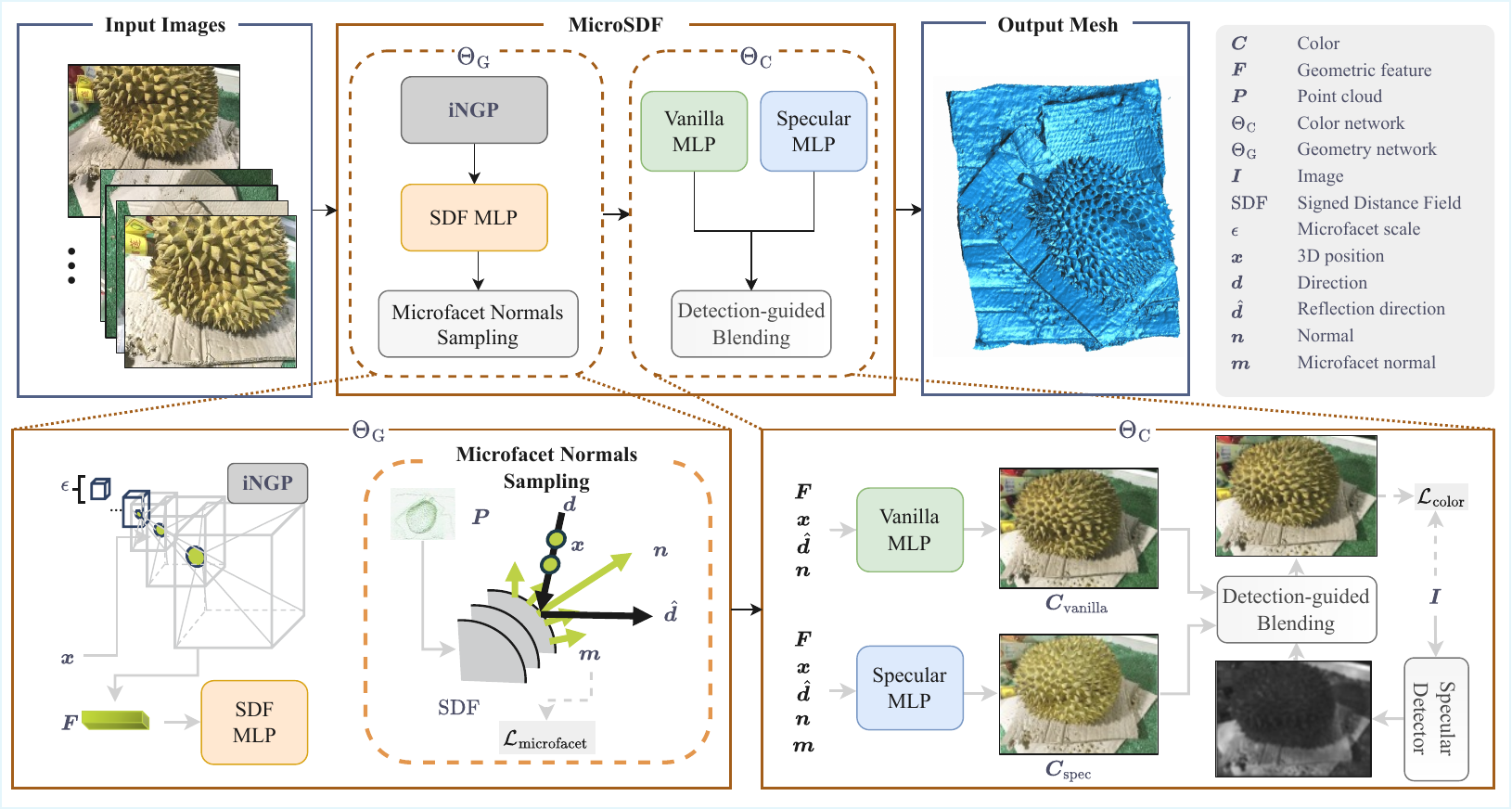}
		\caption{Overview of the proposed MicroSDF. It comprises a geometry network $\Theta_{\text{G}}$ and a color network $\Theta_{\text{C}}$. The geometry network samples spatial points $\boldsymbol{x}$, encodes them via an instant Neural Graphics Primitive(iNGP)~\cite{muller2022instant} to obtain features $\boldsymbol{F}$, and passes these to the SDF MLP, which is initialized using the external point cloud $\boldsymbol{P}$. The color network $\Theta_{\text{C}}$ consists of Vanilla MLP and Specular MLP, both receiving the same inputs: geometric features $\boldsymbol{F}$, points $\boldsymbol{x}$, reflection direction $\hat{\boldsymbol{d}}$ (computed from normal $\boldsymbol{n}$ and viewing direction $\boldsymbol{d}$), and normal $\boldsymbol{n}$. The Specular MLP additionally takes microfacet normals $\boldsymbol{m}$ as input. The outputs from the two branches are fused via the proposed detection-guided blending module, where the blending weights are predicted by an additional specular detector, yielding the final color.}
		\label{fig:Network Architecture}
		\vspace{-0.5cm}
	\end{figure*}
	
	\section{The Proposed MicroSDF}
	\label{The-Proposed-MicroSDF}
	
	In this section, we introduce the proposed MicroSDF, a unified neural surface reconstruction framework that integrates microfacet theory to jointly model geometry and reflectance for both reflective and non-reflective surfaces. An overview of the proposed framework is illustrated in~Fig.~\ref{fig:Network Architecture}. 
	This section elaborates the overall design of our method, beginning with rendering preliminaries (Section~\ref{sec:Preliminaries}), followed by two core components: (\textbf{i}) microfacet-guided geometry modeling (Section~\ref{sec:geometry_learning}), which extracts physically consistent surface normals from a signed distance field, and (\textbf{ii}) microfacet-driven color modeling (Section~\ref{sec:Microfacet-driven dual-branch color modeling}), which employs a dual-branch network with a proposed detection-guided blending process. We also describe our multi-stage optimization strategy (Section~\ref{sec:optimization}), which enables progressive refinement and ensures physically faithful reconstruction.
	
	\vspace{-0.2cm}
	\subsection{Preliminaries}
	\label{sec:Preliminaries}
	
	\subsubsection{Volume Rendering in Neural Radiance Fields}
	Neural Radiance Fields (NeRF) represent a scene by mapping a 3D position $\boldsymbol{x}$ and view direction $\boldsymbol{d}$ to density $\sigma$ and color $\boldsymbol{c}$ via MLPs. For each camera ray, points are sampled along its path, and the final pixel color is computed by volume rendering:
	\begin{equation}
		\boldsymbol{C} = \sum_{i} w_i \boldsymbol{c}_i, \quad w_i = T_i \alpha_i,
	\end{equation}
	where $\alpha_i = 1 - \exp(-\sigma_i \delta_i)$ is the opacity, $\delta_i$ is the interval between samples, and $T_i = \prod_{j=1}^{i-1} (1 - \alpha_j)$ is the accumulated transmittance. The network is trained by minimizing the color reconstruction loss:
	\begin{equation}
		\mathcal{L}_{\text{color}} = \sum \|\boldsymbol{C} - \boldsymbol{C}_{\text{gt}}\|^2,
	\end{equation}
	where $\boldsymbol{C}_{\text{gt}}$ denotes the ground-truth pixel color.
	
	\subsubsection{Surface Rendering}
	The traditional rendering equation describes the outgoing radiance \(L_o\) from a surface point \(\boldsymbol{x}\) in direction \(\boldsymbol{v}\):
	\begin{equation}
		L_o(\boldsymbol{x}, \boldsymbol{v}) = \int_{\Omega} L_i(\boldsymbol{x}, \boldsymbol{v}, \boldsymbol{l}) \, f_r(\boldsymbol{x}, \boldsymbol{v}, \boldsymbol{l}) \, d\boldsymbol{l} + L_e(\boldsymbol{x}, \boldsymbol{v}),
		\label{eq: rendering-function}
	\end{equation}
	where \(L_i(\boldsymbol{x}, \boldsymbol{v}, \boldsymbol{l})\) is the incoming radiance from direction \(\boldsymbol{l}\), \(f_r(\boldsymbol{x}, \boldsymbol{v}, \boldsymbol{l})\) is the Bidirectional Reflectance Distribution Function (BRDF), \(L_e(\boldsymbol{x}, \boldsymbol{v})\) is the emitted light, and the integral is over the hemisphere \(\Omega\). 
	
	Cook-Torrance\cite{cook1982reflectance} model was established based on microfacet theory, assuming surfaces consist of tiny mirror-like facets. 
	Its specular BRDF at point \(\boldsymbol{x}\) is:
	\begin{equation}
		f_{\text{cook-torrance}}(\boldsymbol{x}, \boldsymbol{\omega_o}, \boldsymbol{\omega_i}) = \frac{D(\boldsymbol{x}) \cdot F(\boldsymbol{x}, \boldsymbol{\omega_o}) \cdot G(\boldsymbol{x}, \boldsymbol{\omega_o}, \boldsymbol{\omega_i})}{4 \cdot (\boldsymbol{\omega_o} \cdot \boldsymbol{n}) \cdot (\boldsymbol{\omega_i} \cdot\boldsymbol{n} )},
		\label{eq:render-eq}
	\end{equation}
	where \(D(\boldsymbol{x})\) is the normal distribution function, \(F(\boldsymbol{x}, \boldsymbol{\omega_o})\) is the Fresnel function, \(G(\boldsymbol{x}, \boldsymbol{\omega_o}, \boldsymbol{\omega_i})\) is the geometry function, \(\boldsymbol{\omega_o}\) and \(\boldsymbol{\omega_i}\) are the observation and incident directions, and \(\boldsymbol{n}\) is the surface normal at \(\boldsymbol{x}\).
	
	\vspace{-0.2cm}
	\subsection{Microfacet-Guided Geometry Modeling}
	\label{sec:geometry_learning}
	\subsubsection{Geometry Network}
	The geometry network, parameterized by $\boldsymbol{\Theta}_G$, is designed for capturing the underlying 3D structure of the scene. $\boldsymbol{\Theta}_G$ takes as input the sampled 3D coordinates $\boldsymbol{x}$ and the external point cloud $\boldsymbol{P}$, and outputs the corresponding surface normal vectors $\boldsymbol{n}$, geometric features $\boldsymbol{F}$, and SDF values. 
	To encode the spatial properties of the scene,
	the 3D coordinates $\boldsymbol{x}$ are first fed into an iNGP, which produces geometric features $\boldsymbol{F}$. These features are then further passed through an MLP to predict the SDF value. To facilitate effective training, especially in the early stages, the external point cloud $\boldsymbol{P}$ is used to directly supervise the SDF predictions.
	The resulting outputs provide essential geometric context for the subsequent color modeling stage.
	
	\subsubsection{Dual-Scale Normal Extraction}
	\label{sec:microfacet_normals_sampling}
	
	Accurately modeling surface reflectance, especially specular components, requires not only capturing the overall surface shape but also accounting for fine-grained normal variations. To this end, we propose to extract dual-scale normals that compute both macroscopic and microfacet normals for each sampled 3D point.
	This dual-scale formulation enables our model to capture both global surface structure and fine-grained normal variations, which are critical for consistent geometry reconstruction and physically plausible specular reflection modeling. 
	
	Given a sampled 3D point \(\boldsymbol{x}_i\), two types of normals are computed as follows:
	(\textbf{i}) \textbf{Macroscopic normals} \(\boldsymbol{n}\) are estimated via central differences along coordinate axes with a decreasing step size $\epsilon$ during training. For example, the x-directional gradient is computed as:
	\begin{equation}
		\nabla_x f(\boldsymbol{x}_i) = \frac{\Phi(\boldsymbol{x}_i + \epsilon_{\boldsymbol{x}}) - \Phi(\boldsymbol{x}_i - \epsilon_{\boldsymbol{x}})}{2\epsilon},
	\end{equation}
	where $\epsilon_{\boldsymbol{x}} = [\epsilon, 0, 0]$, and \(\Phi(\boldsymbol{x})\) is the SDF learned by the geometry network, representing the distance from point \(\boldsymbol{x}\) to the surface.
	(\textbf{ii}) \textbf{Microfacet normals} \(\boldsymbol{m}_{i}\) are computed at a finer scale (eg. level \(L+4\) when training at level \(L\)), using the first-order derivative of \(\Phi(\boldsymbol{x})\) as an analytical solution. 
	The sampled macroscopic normal \(\boldsymbol{n}\) and microfacet normals \(\boldsymbol{m}_i\) are concatenated to form a multi-scale normal representation, which is later used to predict the specular color component in our color modeling network (Section~\ref{Dual-branch color network}).  
	
	This hierarchical sampling strategy is grounded in microfacet theory, where surfaces are modeled as collections of tiny mirror-like facets. By leveraging this representation, we can better capture the directional behavior of specular reflections. 
	Note that, unlike AniSDF\cite{gao2024anisdf}, which requires maintaining two separate geometric networks at different granularities, our approach achieves multi-scale learning within a single geometry network, enabling more efficient and physically consistent training.
	
	\begin{figure}[t]
		\centering
		\includegraphics[width=1.0\linewidth]{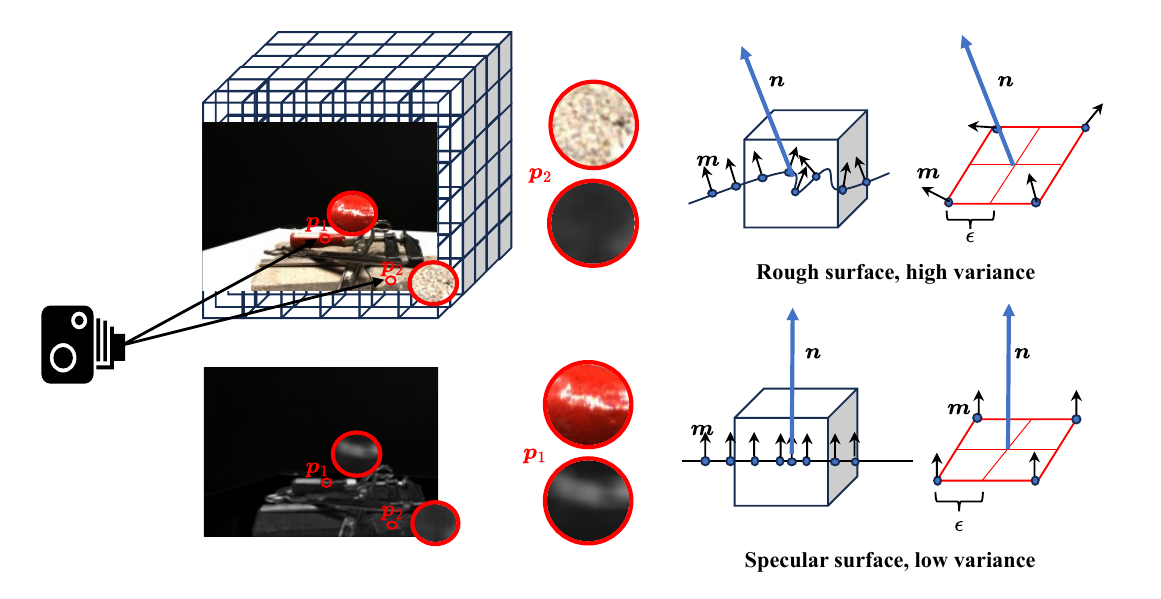}
		\vspace{-0.5cm}
		\caption{Microfacet normal characteristics. $\boldsymbol{p_1}$ and $\boldsymbol{p_2}$ illustrate representative points on specular and diffuse surfaces, respectively. According to microfacet theory, the normal variance at specular surfaces ($\boldsymbol{p_1}$) is smaller than that at diffuse surfaces ($\boldsymbol{p_2}$).}
		\label{fig:microfacet_normals}
		\vspace{-0.5cm}
	\end{figure}
	
	\subsubsection{Microfacet Normal Consistency}
	As illustrated in 
	\label{sec:microfacet normal consistency}
	Fig.~\ref{fig:microfacet_normals}, specular regions (\textit{e.g.}, $\boldsymbol{p_1}$) tend to exhibit low variance among microfacet normals, while rough regions (\textit{e.g.}, $\boldsymbol{p_2}$) show higher variance, consistent with classical microfacet theory. 
	To enforce alignment between macroscopic and microscopic surface orientations, we define a consistency metric based on the angular variance:
	\vspace{-0.2cm}
	\begin{equation}
		\tau^2 = \frac{1}{N} \sum_{i=1}^{N} \left(1 - \boldsymbol{n} \cdot \boldsymbol{m}_i \right),
	\end{equation}
	where \( N \) denotes the number of microfacet normals.
	This measures the average angular discrepancy between the coarse surface normal $\boldsymbol{n}$ and each microfacet normal $\boldsymbol{m}_i$. The microfacet consistency loss is then defined as:
	\vspace{-0.2cm}
	\begin{equation}
		\mathcal{L}_{\text{microfacet}} = \frac{1}{M} \sum_{i=1}^{M} \left( \tau^2_i \cdot mask_i \cdot spec_i \right), \label{microfacet-loss}
	\end{equation}
	where $M$ denotes the number of sampled points, \(mask_i\) indicates valid surface regions, \(spec_i\) is the specular score from the detection network (Section~\ref{sec:Specular-Detection}). 
	This loss enforces the physically grounded constraint that specular regions should exhibit consistent microfacet orientations, thereby regularizing the geometry network to produce normals that are directionally coherent with specular reflection behavior.
	
	\subsection{Microfacet-Driven Color Modeling}
	\label{sec:Microfacet-driven dual-branch color modeling}

	\subsubsection{Dual-branch color network}
	\label{Dual-branch color network}
	
	We decompose the color modeling process into two branches: (\textbf{i}) \textbf{Vanilla Branch:} Following Ref-NeRF~\cite{2022refnerf}, we first compute the reflection direction \(\hat{\boldsymbol{d}}\) from the normal vector \(\boldsymbol{n}\) and the ray direction \(\boldsymbol{d}\). 
	The Vanilla MLP takes as inputs \(\boldsymbol{x}\), \(\hat{\boldsymbol{d}}\), \(\boldsymbol{n}\), and \(\boldsymbol{F}\), and outputs the vanilla color 
	$\boldsymbol{C}_{vanilla}$, formulated as:
	\begin{equation}
		\boldsymbol{C}_{vanilla} = \text{MLP}_{\text{vanilla}}(\boldsymbol{x}, \hat{\boldsymbol{d}}, \boldsymbol{n}, \boldsymbol{F}).
	\end{equation}
	As evidenced by Ref-NeRF\cite{2022refnerf}, the reflection direction $\hat{\boldsymbol{d}}$ is adequate for modeling most real-world scenarios, allowing the vanilla branch to absorb low-frequency specular cues and reducing the burden on the specular branch. 
	Our ablation studies (see Fig.~\ref {fig:ablation_view_dir_vs_ref_dir} and Table~\ref {tab:ablation-combined}) further demonstrate the advantages of this design.
	(\textbf{ii}) \textbf{Specular Branch:} To more accurately model specular reflections, we augment the input of the Specular MLP with microfacet normals \(\boldsymbol{m}\), sampled as described in Section~\ref{sec:microfacet_normals_sampling}. This branch also receives (\(\boldsymbol{x}\), \(\hat{\boldsymbol{d}}\), \(\boldsymbol{n}\), and \(\boldsymbol{F}\)), leveraging both macroscopic and fine-grained geometric cues, and outputs the specular color component for each sampled point, formulated as:
	\begin{equation}
		\boldsymbol{C}_{\text{specular}} = \text{MLP}_{\text{specular}}(\boldsymbol{x}, \hat{\boldsymbol{d}},\boldsymbol{n} ,\boldsymbol{m} ,\boldsymbol{ F}).
	\end{equation}
	Introducing microfacet normals $\boldsymbol{m}$  enables us to empirically estimate \(D(\boldsymbol{x})\) in Eq.~\ref{eq:render-eq}, improving the network's ability to reconstruct specular regions in a physically grounded manner.

	\subsubsection{Detection-guided blending}
	\label{sec:Specular-Detection}
	Handling reflective and non-reflective surfaces within a unified framework requires not only accurate modeling of each component but also precise spatial allocation of where these components should dominate. A key challenge in dual-branch color networks lies in ambiguity when decomposing color into low-frequency and high-frequency specular components.
	
	To address this, we propose a detection-guided color blending mechanism that leverages external, physically interpretable priors to control the blending between the vanilla branch (responsible for low-frequency specular effects) and the specular branch (capturing high-frequency specular highlights).
	Unlike UniSDF\cite{wang2023unisdf} and AniSDF\cite{gao2024anisdf}, which infer blending weights via an auxiliary MLP, making the fusion process susceptible to dataset bias and prone to misidentifying specular cues,
	we employ StableDelight~\cite{stableDelight2025}, a diffusion-based specular detector, to per-pixel specular confidence maps directly from the input images. 
	These detection scores encode the likelihood of each pixel exhibiting high-frequency specular reflection, thus providing explicit, spatially varying blending weights grounded in physical intuition.
	The final pixel color is computed as:
	\begin{equation}
		\boldsymbol{C} = (1-\text{Spec}) \cdot \boldsymbol{C}_{vanilla} + \text{Spec} \cdot \boldsymbol{C}_{specular}. 
	\end{equation}
	where $\text{Spec}$ denotes specular score. 
	
	\subsection{Optimization}
	\label{sec:optimization}
	Our optimization process consists of three key components: geometry initialization, a multi-stage training strategy, and our loss function design.

	\subsubsection{Geometry Initialization}
	\label{sec:Geometry Initialization}
	To provide a reliable starting point for the geometry network \(\boldsymbol{\Theta}_G\), we initialize the SDF \(\Phi(\boldsymbol{x})\) using dense point cloud information obtained from traditional 3D reconstruction algorithms (\textit{e.g.}, OpenMVS~\cite{openmvs_github}). These point clouds help constrain surface positions and identify empty regions in 3D space. 
	
	Inspired by the regularization approach in SIREN~\cite{sitzmann2020implicit}, we encourage the SDF values of surface points to approach 0. 
	However, unlike SIREN~\cite{sitzmann2020implicit}, which incorporates normal constraints, our microfacet-based rendering method is particularly sensitive to inaccuracies in surface normals. To avoid the detrimental effects of the initial normal errors, we exclude normal constraints during the geometry initialization stage.
	Besides, we randomly sample off-surface points and enforce that their SDF gradients are close to zero. This encourages flatness in empty regions and contributes to a more stable and robust initialization. The specific loss function is defined as:
	\vspace{-0.2cm}
	\begin{equation}
		\begin{split}
			\mathcal{L}_{\mathrm{sdf}} = & \int_{\Omega} \left| \left\| \boldsymbol{\nabla}_{\boldsymbol{x}} \Phi(\boldsymbol{x}) \right\| - 1 \right| \, d\boldsymbol{x} \\
			&+ \int_{\Omega_0} \left\| \Phi(\boldsymbol{x}) \right\| \, d\boldsymbol{x} + \int_{\Omega \setminus \Omega_0} \psi \left( \Phi(\boldsymbol{x}) \right) \, d\boldsymbol{x},
		\end{split}
	\end{equation}
	where \(\Omega\) denotes the entire 3D space, \(\Omega_0\) is the surface region of the point cloud, and \(\psi\)() is a penalty function applied to off-surface SDF values. The first term ensures that the SDF satisfies the eikonal equation, the second term enforces consistency of surface positions, and the third term penalizes deviations in off-surface points.

	To further distinguish occupied from empty regions, we utilize foreground masks of input images by leveraging established fast matting algorithms, such as \cite{lee2022tracer}. We enforce the accumulated weight of sampled points along each background ray path to be minimized. This mask loss is defined as:
	\begin{equation}
		\mathcal{L}_{\text{mask}} = \left\| \boldsymbol{GT}_{\text{mask}} - \int_{t_n}^{t_f} \boldsymbol{T}(t) \cdot \sigma(t) \, dt \right\|^2,
	\end{equation}
	where \(\boldsymbol{GT}_{\text{mask}}\) is the ground-truth mask extracted from the input images, \(\boldsymbol{T}(t)\) and \(\sigma(t)\) are the transmittance and density along the ray \(t\), respectively, and \(t_n\) and \(t_f\) represent the near and far boundaries of the ray.
	
	\subsubsection{Four-stage Training Strategy}
	\label{sec:training-strategy}
	Although the effectiveness of coarse-to-fine training has been well demonstrated, our experiments show that, with sufficient training at the initial level, certain intermediate levels can be skipped. This enables a more direct transition to the finest level, where analytical gradients can replace numerical ones for improved efficiency.
	Based on this insight, we propose a four-stage training strategy comprising: (\textbf{i}) geometry initialization (Geometry Init.), (\textbf{ii}) thorough training at the first level (1st Level), (\textbf{iii}) progressive coarse-to-fine processing (Coarse-to-Fine), and (\textbf{iv}) final training at the finest level (Finest Level). 
	
	This training strategy significantly reduces the total number of training iterations compared to conventional coarse-to-fine training approaches, while yielding superior quantitative results (see Table VI in Supplementary Material). Our microfacet sampling process is mainly performed during the thorough training at the first level and the progressive coarse-to-fine processing stages.

	\subsubsection{Loss function design}
	\label{sec:basic-losses}
	Our loss function design is closely aligned with the four-stage training strategy, with specific loss terms activated at different training stages.  A summary of the loss terms at each stage is provided in Table~\ref{tab:losses-per-stage}.
	
	\begin{table}[t]
		\centering
		\caption{Loss terms used at different training stages.}
		\label{tab:losses-per-stage}
		\footnotesize
		\begin{tabular}{lcccc}
			\toprule
			& Geometry Init. & 1st Level & Coarse-to-Fine & Finest Level \\
			\midrule
			$\mathcal{L}_{\mathrm{sdf}}$ & \checkmark & & & \\
			$\mathcal{L}_{\mathrm{mask}}$ & \checkmark & & & \\
			$\mathcal{L}_{\text{color}}$ & & \checkmark & \checkmark & \checkmark \\
			$\mathcal{L}_{\text{eik}}$ & & \checkmark & \checkmark & \checkmark \\
			$\mathcal{L}_{\text{curv}}$ & & \checkmark & \checkmark & \\
			$\mathcal{L}_{\text{microfacet}}$ & & \checkmark & \checkmark & \\
			\bottomrule
		\end{tabular}
		\vspace{-0.5cm}
	\end{table}

	As the geometry initialization stage (detailed in Section~\ref{sec:Geometry Initialization}) focuses on constraining surface geometry and identifying empty space, we define a composite loss function:
	\begin{equation}
		\mathcal{L}_{\mathrm{Geometry \ Init.}} = \lambda_{\mathrm{sdf}} \mathcal{L}_{\mathrm{sdf}} + \lambda_{\mathrm{mask}} \mathcal{L}_{\mathrm{mask}}
	\end{equation}
	where \(\lambda_{\mathrm{sdf}}\) and \(\lambda_{\mathrm{mask}}\) are weighting coefficients balancing geometric fidelity and empty region constraints.
	In the 1st Level and Coarse-to-Fine stages, the primary objective shifts to jointly refining geometry and color. The core loss components include:
	(\textbf{i}) \textbf{Color loss}: This loss minimizes the discrepancy between the rendered color and the input image pixel color:
	\begin{equation}
		\mathcal{L}_{\text{color}} = \sum{\left\| \boldsymbol{C} - \boldsymbol{C}_{\text{gt}} \right\|^2}
	\end{equation}
	where \( C \) denotes the rendered color, and \( C_{\text{gt}} \) represents the ground-truth color.
	(\textbf{ii}) \textbf{Eikonal loss}~\cite{gropp2020implicit}: To maintain the validity of the SDF features, we enforce the Eikonal constraint:
	\begin{equation}
		\mathcal{L}_{\text{eik}} = \sum{ \left( \left\| \nabla \Phi(\boldsymbol{x}) \right\| - 1 \right)^2 }
	\end{equation}
	(\textbf{iii}) \textbf{Curvature loss}: Following Neuralangelo~\cite{li2023neuralangelo}, we introduce a curvature regularization based on the second-order derivatives of the SDF, which promotes smooth surface reconstructions:
	\begin{equation}
		\mathcal{L}_{\text{curv}} =  \sum{\left( \left\| \nabla^2 \Phi({x}_i) \right\| \right)}
	\end{equation}
	where \(\nabla^2 \Phi(\boldsymbol{x}_i)\) denotes the  Laplacian of the SDF value at the sampled point \(\boldsymbol{x}_i\), reflecting local curvature.
	(\textbf{iv}) \textbf{Microfacet consistency loss}: To enhance the fidelity of reflectance modeling, we introduce a microfacet consistency loss that aligns predicted surface orientations with the microfacet-based reflectance model (Detailed in Eq.~\ref{microfacet-loss} in Section~\ref{sec:microfacet normal consistency}).
	The combined loss during the second and third stages is:
	\begin{align}
		\mathcal{L}_{\mathrm{1st\ Level\ \&\ Coarse\text{-}to\text{-}Fine}} =\ 
		&\mathcal{L}_{\text{color}} \notag + \lambda_{\text{eik}}\, \mathcal{L}_{\text{eik}} \notag + \lambda_{\text{curv}}\, \mathcal{L}_{\text{curv}} \notag \\
		&+ \lambda_{\text{microfacet}}\, \mathcal{L}_{\text{microfacet}} 
	\end{align}
	where \(\lambda_{\text{eik}}\), \(\lambda_{\text{curv}}\), and \(\lambda_\text{microfacet}\) are weighting factors of their respective loss terms. 
	In the final stage, analytical gradients are used, making curvature and microfacet losses unnecessary. The objective loss function is:
	\begin{equation}
		\mathcal{L}_{\mathrm{Finest \ Level}} = \mathcal{L}_{\text{color}} + \lambda_{\text{eik}}\, \mathcal{L}_{\text{eik}}.
	\end{equation}
	\begin{figure*}[t]
		\centering
		\includegraphics[width=0.95\linewidth]{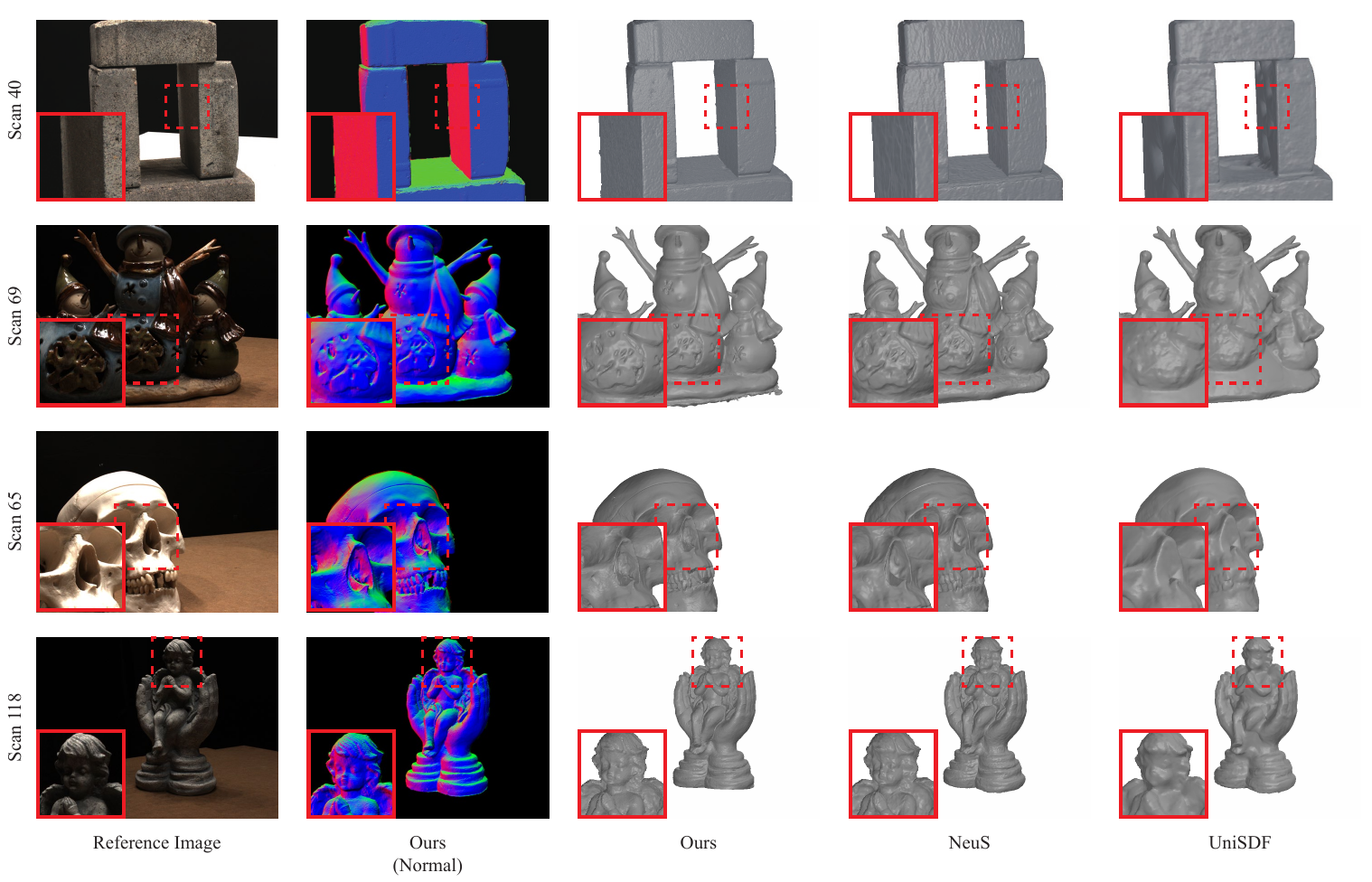}
		\vspace{-0.5cm}
		\caption{Qualitative comparison results on the DTU benchmark. Our approach achieves reconstruction that preserves finer geometric details compared to the state-of-the-art neural implicit method, NeuS~\cite{NEURIPS2021_e41e164f}, while significantly outperforming the general-purpose reconstruction method UniSDF~\cite{wang2023unisdf}.}
		\label{fig:dtu-results}
		\vspace{-0.2cm}
	\end{figure*}
	\begin{figure*}
		\centering
		\includegraphics[width=0.95\linewidth]{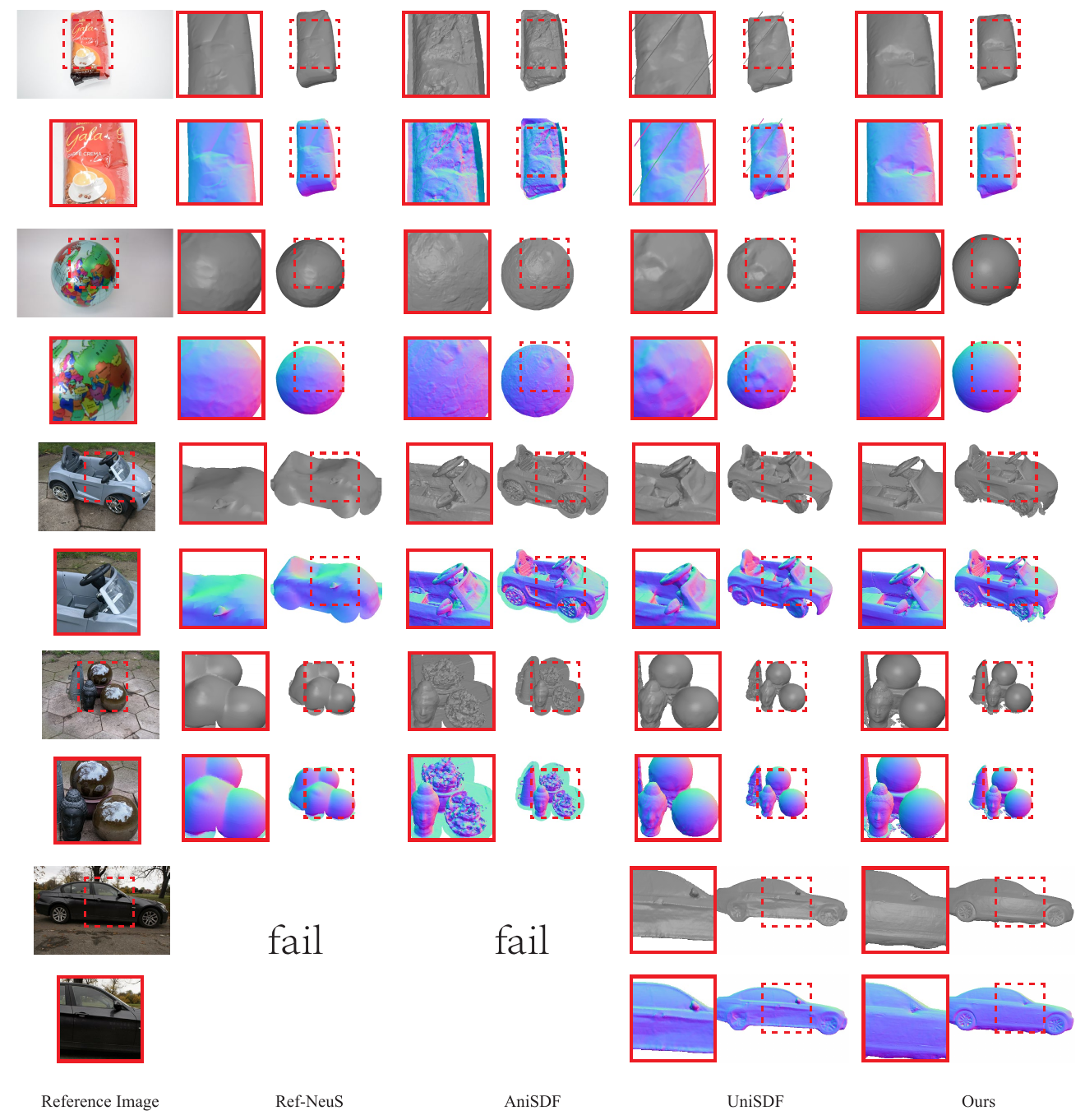} 
		\vspace{-0.5cm}
		\caption{Qualitative comparison of the reflective surface reconstruction between the state-of-the-art methods and ours on the DeepVoxels (top two scenes) and Ref-NeRF (bottom three scenes) datasets. Our method outperforms UniSDF\cite{wang2023unisdf} and AniSDF\cite{gao2024anisdf} with higher success rates and better geometric fidelity.}
		\label{fig:ref-nerf-results}
		\vspace{-0.5cm}
	\end{figure*}
	
	\begin{table*}[t]
		\caption{Reconstruction performance on the DTU dataset in terms of Chamfer distance(mm) $\downarrow$ (lower is better). \colorbox{red!25}{Red} and \colorbox{orange!25}{orange} denote the best and the second-best results, respectively. Results marked with * are sourced from the UniSDF~\cite{wang2023unisdf} paper, while others are reported from their respective papers.}
		\label{tab:dtu-results}
		\vspace{-0.2cm}
		\centering
		\begin{tabularx}{\textwidth}{l *{17}{>{\centering\arraybackslash}X}}
			\toprule
			Methods &  24 & 37 & 40 & 55 & 63 & 65 & 69 & 83 & 97 & 105 & 106 & 110 & 114 & 118 & 122 & Mean & Runtime\\
			\midrule
			VolSDF*~\cite{yariv2021volume} & 1.14 & 1.26 & 0.81 & 0.49 & 1.25 & 0.70 & 0.72 & 1.29 & 1.18 & 0.70 & 0.66 & 1.08 & 0.42 & 0.61 & 0.55 & 0.86 &
			\mbox{$~\approx 8h$}\\
			NeuS*~\cite{NEURIPS2021_e41e164f} & 1.37 & 1.21 & 0.73 & 0.40 & 1.20 & 0.70 & 0.72 & 1.01 & 1.16 & 0.82 & 0.66 & 1.69 & 0.39 & \cellcolor{orange!25}0.49 & 0.51 & 0.87 & \mbox{$~\approx 8h$} \\
			NeuralWarp*~\cite{darmon2022improving} & 0.49 & \cellcolor{orange!25}0.71 & 0.38 & \cellcolor{orange!25}0.38 & \cellcolor{orange!25}0.79 & 0.81 & 0.82 & 1.20 & 1.06 & 0.68 & 0.66 & 0.74 & 0.41 & 0.63 & 0.51 & 0.68 & \mbox{$~\approx 8h$}\\
			Neuralangelo*~\cite{li2023neuralangelo} & 0.49 & 1.05 & 0.95 & \cellcolor{orange!25}0.38 & 1.22 & 1.10 & 2.16 & 1.68 & 1.78 & 0.93 & \cellcolor{red!25}0.44 & 1.46 & 0.41 & 1.13 & 0.97 & 1.08 & \mbox{$~\approx 16h$} \\
			PermutoSDF*~\cite{rosu2023permutosdf} &  0.52 & 0.75 & 0.41 & \cellcolor{red!25}0.37 & 0.90 & 0.66 & \cellcolor{orange!25}0.59 & 1.37 & 1.07 & 0.85 & \cellcolor{orange!25}0.46 & 0.98 & \cellcolor{red!25}0.33 & \cellcolor{red!25}0.39 & 0.50 & 0.68 & \mbox{$~\approx 1h$} \\
			UniSDF*~\cite{wang2023unisdf} & 0.54 & 0.84 & 0.66 & 0.51 & \cellcolor{red!25}0.76 & \cellcolor{orange!25}0.64 & 0.71 & \cellcolor{red!25}0.70 & \cellcolor{red!25}0.86 & \cellcolor{orange!25}0.57 & 0.69 & \cellcolor{red!25}0.65 & 0.45 & 0.56 & 0.50 & \cellcolor{orange!25}0.64 & \mbox{$~\approx 1h$}\\
			MonoSDF~\cite{yu2022monosdf} & 0.66 & 0.88 & 0.43 & 0.40 & 0.87 & 0.78 & 0.81 & 1.23 & 1.18 & 0.66 & 0.66 & 0.96 & 0.41 & 0.57 & 0.51 & 0.73 & \mbox{$~\approx 8h$} \\
			AniSDF~\cite{gao2024anisdf} & 0.52 & 0.82 & 0.65 & 0.43 & \cellcolor{red!25}0.76 & \cellcolor{orange!25}0.64 & 0.71 & \cellcolor{orange!25}0.97 & \cellcolor{red!25}0.86 & 0.64 & 0.52 & 0.67 & 0.42 & 0.67 & 0.50 & 0.65 & \mbox{$~\approx 1h$}\\
			GSDF~\cite{yu2024gsdf} & 0.59 & 0.94 & 0.46 & \cellcolor{orange!25}0.38 & 1.30 &  0.77 & 0.73 &  1.59 &  1.29 & 0.76 &  0.59 &  1.22 &  0.38 &  0.52 &  0.51 & 0.80 &  \mbox{$~\approx 8h$}\\
			SuGaR~\cite{guedon2024sugar} & 1.47 & 1.33 & 1.13 & 0.61 & 2.25 & 1.71 & 1.15 & 1.63 & 1.62 & 1.07 & 0.79 & 2.45 & 0.98 & 0.88 & 0.79 & 1.33 & \mbox{$~\approx 1h$}\\
			2DGS~\cite{huang20242d} & \cellcolor{orange!25}0.48 & 0.91 & 0.39 & 0.39 & 1.01 & 0.83 & 0.81 & 1.36 & 1.27 & 0.76 & 0.70 & 1.40 & 0.40 & 0.76 & 0.52 & 0.80 & \mbox{$~\approx 0.5h$}\\
			GOF~\cite{yu2024gaussian} & 0.50 & 0.82 & 0.37 & \cellcolor{red!25}0.37 & 1.12 & 0.74 & 0.73 & 1.18 & 1.29 & 0.68 & 0.77 & 0.90 & 0.42 & 0.66 & 0.49 & 0.74 & \mbox{$~\approx 1h$}\\
			Rade-GS~\cite{zhang2024rade} & \cellcolor{red!25}0.46 & 0.73 & \cellcolor{orange!25}0.33 & \cellcolor{orange!25}0.38 & \cellcolor{orange!25}0.79 & 0.75 & 0.76 & 1.19 & 1.22 & 0.62 & 0.70 & 0.78 & \cellcolor{orange!25}0.36 & 0.68 & \cellcolor{orange!25}0.47 & 0.68 & \mbox{$~\approx 0.5h$}\\
			Ours & 0.53 & \cellcolor{red!25}0.64 & \cellcolor{red!25}0.31 & 0.44 & 0.88 & \cellcolor{red!25}0.62 & \cellcolor{red!25}0.56 & 1.34 & \cellcolor{orange!25}1.00 & \cellcolor{red!25}0.52 & 0.48 & \cellcolor{orange!25}0.66 & 0.39 & \cellcolor{red!25}0.39 & \cellcolor{red!25}0.40 & \cellcolor{red!25}0.61 & \mbox{$~\approx 1h$}\\
			\bottomrule
		\end{tabularx}
		\vspace{-0.5cm}
	\end{table*}

	\begin{table*}[t]
		\centering
		\caption{Quantitative results on Shiny Blender dataset. Lower is better for both MAE$^\circ$ and ACC. \colorbox{red!25}{Red} and \colorbox{orange!25}{orange} denote the best and the second-best results, respectively. Results marked with * are sourced from the UniSDF~\cite{wang2023unisdf} paper, while others are reported from their respective papers.}
		\label{tab:shiny_blender}
		\small
		\begin{tabular}{l*{7}{c}*{5}{c}}
			\toprule
			\multirow{2}{*}{Method} & \multicolumn{7}{c}{MAE$^\circ$ $\downarrow$} & \multicolumn{5}{c}{ACC $\downarrow$} \\
			\cmidrule(lr){2-8} \cmidrule(lr){9-13}
			& car & ball & helmet & teapot & toaster & coffee & mean & car & helmet & toaster & coffee & mean \\
			\midrule
			Geo-NeuS*~\cite{fu2022geo} & 12.15 & 1.04 & 4.12 & 19.77 & 16.23 & 9.79 & 10.52 & 0.72 & 0.74 & 4.14 & \cellcolor{orange!25}0.90 & 1.63 \\
			Neuralangelo*~\cite{li2023neuralangelo} & 12.34 & 35.63 & 9.23 & 4.98 & 14.14 & 8.61 & 14.16 & 1.89 & 1.71 & 2.99 & \cellcolor{red!25}0.63 & 1.81 \\
			Ref-NeRF*~\cite{2022refnerf} & 14.93 & 1.55 & 29.48 & 9.23 & 42.87 & 12.24 & 18.38 & -- & -- & -- & -- & -- \\
			NMF~\cite{mai2023neural} & 7.74 & 0.72 & 2.40 & 5.67 & \cellcolor{orange!25}6.66 & 13.17 & 6.06 & -- & -- & -- & -- & -- \\
			Ref-NeuS*~\cite{ge2023ref-neus} & 7.68 & \cellcolor{orange!25}0.50 & 1.94 & 10.25 & \cellcolor{red!25}5.95 & \cellcolor{red!25}5.73 & 5.34 & \cellcolor{orange!25}0.50 & 0.53 & \cellcolor{red!25}0.54 & 1.83 & \cellcolor{orange!25}0.85 \\
			UniSDF*~\cite{wang2023unisdf} & \cellcolor{red!25}6.88 & \cellcolor{red!25}0.45 & \cellcolor{red!25}1.72 & \cellcolor{red!25}2.80 & 8.71 & 8.00 & \cellcolor{orange!25}4.76 & 0.58 & \cellcolor{orange!25}0.46 & 2.02 & 1.17 & 1.06 \\
			\textbf{Ours} & \cellcolor{orange!25}7.15 & 0.51 & \cellcolor{orange!25}1.77 & \cellcolor{orange!25}3.60 & 7.92 & \cellcolor{orange!25}6.55 & \cellcolor{red!25}4.58 & \cellcolor{red!25}0.35 & \cellcolor{red!25}0.24 & \cellcolor{orange!25}0.90 & 0.94 & \cellcolor{red!25}0.61 \\
			\bottomrule
		\end{tabular}
	\end{table*}

	\begin{table*}[t]
		\caption{Ablation on microfacet normals sampling (MS), reflective direction of Vanilla MLP (RD), and detection-based color blend weighting strategy (DC). All metrics are reported in Chamfer Distance (mm) $\downarrow$ (lower is better). \colorbox{red!25}{Red} and \colorbox{orange!25}{orange} denote top and second-best results. When DC=$\times$, blending weights are learned via MLP as in UniSDF~\cite{wang2023unisdf}.}
		\label{tab:ablation-combined}
		\vspace{-0.2cm}
		\centering
		\begin{tabularx}{\textwidth}{l l l *{17}{>{\centering\arraybackslash}X}}
			\toprule
			MS & RD & DC & 24 & 37 & 40 & 55 & 63 & 65 & 69 & 83 & 97 & 105 & 106 & 110 & 114 & 118 & 122 & Mean \\
			\midrule
			$\times$ & \checkmark & \checkmark & 0.95 & \cellcolor{red!25}0.62 & \cellcolor{orange!25}0.32 & \cellcolor{orange!25}0.51 & 0.92 & \cellcolor{orange!25}0.58 & \cellcolor{red!25}0.52 & 1.37 & 1.01 & \cellcolor{orange!25}0.51 & 0.50 & 0.69 & 0.35 & 0.43 & \cellcolor{red!25}0.39 & 0.65 \\
			\checkmark & $\times$ & \checkmark & 0.74 & 0.75 & 0.35 & 0.54 & \cellcolor{red!25}0.84 & 0.70 & 0.58 & \cellcolor{red!25}1.33 & \cellcolor{orange!25}0.99 & \cellcolor{red!25}0.50 & 0.50 & \cellcolor{orange!25}0.68 & \cellcolor{orange!25}0.34 & 0.41 & \cellcolor{red!25}0.39 & 0.64 \\
			\checkmark & \checkmark & $\times$ & \cellcolor{orange!25}0.60 & 0.73 & \cellcolor{red!25}0.31 & 0.53 & 0.89 & \cellcolor{red!25}0.51 & \cellcolor{red!25}0.52 & 1.35 & \cellcolor{red!25}0.92 & \cellcolor{red!25}0.50 & \cellcolor{red!25}0.47 & 0.75 & 0.38 & \cellcolor{red!25}0.37 & \cellcolor{red!25}0.39 & \cellcolor{orange!25}0.615 \\
			\checkmark & \checkmark & \checkmark & \cellcolor{red!25}0.53 & \cellcolor{orange!25}0.64 & \cellcolor{red!25}0.31 & \cellcolor{red!25}0.44 & \cellcolor{orange!25}0.88 & 0.62 & \cellcolor{orange!25}0.56 & \cellcolor{orange!25}1.34 & 1.00 & 0.52 & \cellcolor{orange!25}0.48 & \cellcolor{red!25}0.66 & \cellcolor{red!25}0.33 & \cellcolor{orange!25}0.39 & \cellcolor{orange!25}0.40 & \cellcolor{red!25}0.607 \\
			\bottomrule
		\end{tabularx}
		\vspace{-0.5cm}
	\end{table*}

	\begin{table*}[t]
		\centering
		\caption{Ablation study on detection-based color blending (DC) strategies on Shiny Blender dataset. Lower is better for both MAE$^\circ$ and ACC. \colorbox{red!25}{Red} and \colorbox{orange!25}{orange} denote the best and the second-best results, respectively. \textbf{Stable}: StableDelight detection scores; \textbf{All Ones}: all blending weights 1 (fully specular branch); \textbf{All Zeros}: all 0 (fully diffuse branch); \textbf{MLP}: weights learned via auxiliary MLP as in UniSDF~\cite{wang2023unisdf}.}
		\label{tab:shiny_blender_dc}
		\small
		\begin{tabular}{l*{7}{c}*{5}{c}}
			\toprule
			\multirow{2}{*}{DC Config} & \multicolumn{7}{c}{MAE$^\circ$ $\downarrow$} & \multicolumn{5}{c}{ACC $\downarrow$} \\
			\cmidrule(lr){2-8} \cmidrule(lr){9-13}
			& car & ball & helmet & teapot & toaster & coffee & mean & car & helmet & toaster & coffee & mean \\
			\midrule
			Stable & 7.60 & \cellcolor{orange!25}0.54 & 1.84 & 3.44 & 9.88 & 8.92 & 5.37 & 0.44 & 0.29 & 1.37 & 1.10 & 0.80 \\
			All Zeros & 7.44 & \cellcolor{orange!25}0.54 & 1.84 & \cellcolor{orange!25}3.30 & 8.38 & 7.28 & 4.80 & \cellcolor{orange!25}{0.42} & 0.29 & 1.26 & \cellcolor{red!25}0.85 & 0.71 \\
			MLP & \cellcolor{red!25}7.09 & 0.59 & \cellcolor{red!25}1.71 & \cellcolor{red!25}3.16 & \cellcolor{orange!25}8.25 & \cellcolor{orange!25}6.76 & \cellcolor{orange!25}4.59 & \cellcolor{red!25}0.35 & \cellcolor{red!25}0.23 & \cellcolor{orange!25}1.16 & \cellcolor{orange!25}0.86 & \cellcolor{orange!25}0.65 \\
			All Ones & \cellcolor{orange!25}7.15 & \cellcolor{red!25}0.51 & \cellcolor{orange!25}1.77 & 3.60 & \cellcolor{red!25}7.92 & \cellcolor{red!25}6.55 & \cellcolor{red!25}4.58 & \cellcolor{red!25}0.35 & \cellcolor{orange!25}0.24 & \cellcolor{red!25}0.90 & 0.94 & \cellcolor{red!25}0.61 \\
			\bottomrule
		\end{tabular}
	\end{table*}

	\begin{table*}[t]
		\centering
		\caption{Ablation study on detection-based color blending (DC) strategies on DTU dataset (Chamfer Distance (mm) $\downarrow$, lower is better). \colorbox{red!25}{Red} and \colorbox{orange!25}{orange} denote the best and second-best per column. \textbf{Stable}: StableDelight detection scores; \textbf{All Ones}: all blending weights set to 1 (fully using the specular branch); \textbf{All Zeros}: all blending weights set to 0 (fully using the diffuse branch); \textbf{MLP}: weights learned via auxiliary MLP as in UniSDF~\cite{wang2023unisdf}.}
		\label{tab:dtu_dc}
		\small
		\begin{tabularx}{\textwidth}{l *{16}{>{\centering\arraybackslash}X}}
			\toprule
			DC Config & 24 & 37 & 40 & 55 & 63 & 65 & 69 & 83 & 97 & 105 & 106 & 110 & 114 & 118 & 122 & Mean \\
			\midrule
			All Ones & \cellcolor{orange!25}0.60 & \cellcolor{orange!25}0.69 & \cellcolor{orange!25}0.37 & 0.59 & \cellcolor{orange!25}0.87 & 0.58 & 0.61 & 1.46 & 1.06 & 0.53 & 0.48 & \cellcolor{orange!25}0.72 & \cellcolor{red!25}0.34 & \cellcolor{orange!25}0.38 & \cellcolor{orange!25}0.40 & 0.65 \\
			All Zeros & 0.64 & 0.73 & \cellcolor{red!25}0.31 & \cellcolor{orange!25}0.51 & \cellcolor{red!25}0.83 & \cellcolor{orange!25}0.54 & \cellcolor{orange!25}0.55 & 1.37 & 1.05 & \cellcolor{orange!25}0.52 & \cellcolor{red!25}0.46 & 0.76 & \cellcolor{orange!25}0.38 & 0.41 & \cellcolor{red!25}0.39 & 0.63 \\
			MLP & \cellcolor{orange!25}0.60 & 0.73 & \cellcolor{red!25}0.31 & 0.53 & 0.89 & \cellcolor{red!25}0.51 & \cellcolor{red!25}0.52 & \cellcolor{orange!25}1.35 & \cellcolor{red!25}0.92 & \cellcolor{red!25}0.50 & \cellcolor{orange!25}0.47 & 0.75 & \cellcolor{orange!25}0.38 & \cellcolor{red!25}0.37 & \cellcolor{red!25}0.39 & \cellcolor{orange!25}0.615 \\
			Stable & \cellcolor{red!25}0.53 & \cellcolor{red!25}0.64 & \cellcolor{red!25}0.31 & \cellcolor{red!25}0.44 & 0.88 & 0.62 & 0.56 & \cellcolor{red!25}1.34 & \cellcolor{orange!25}1.00 & \cellcolor{orange!25}0.52 & 0.48 & \cellcolor{red!25}0.66 & 0.39 & 0.39 & \cellcolor{orange!25}0.40 & \cellcolor{red!25}0.607 \\
			\bottomrule
		\end{tabularx}
		\vspace{-0.5cm}
	\end{table*}

	\section{Experiments}
	\label{sec: experiments}
	\subsection{Implementation Settings}
	\noindent\textbf{Datasets.} Following prior works~\cite{wang2023unisdf,NEURIPS2021_e41e164f,yariv2021volume,li2023neuralangelo, zhang2021physg}, we evaluate the proposed method on three widely used benchmarks:
	(\textbf{i}) \textbf{DTU}\cite{aanaes2016large}: The DTU multi-view dataset contains 15 real-world scenes, each captured from 49 or 64 viewpoints with high-resolution images. Each scene contains objects with diverse materials and complex geometries, including both reflectance and non-reflectance surfaces. DTU provides ground-truth 3D models and camera trajectories, making it a standard benchmark for quantitative evaluation of point cloud reconstruction.(\textbf{ii}) \textbf{Shiny Blender}\cite{2022refnerf}: The Shiny Blender dataset is a synthetic benchmark consisting of 6 reflective objects rendered under controlled settings. We use it to provide additional quantitative evaluation on reflective surfaces.
	(\textbf{iii}) \textbf{Ref-NeRF}\cite{2022refnerf}: The Ref-NeRF dataset is designed to evaluate reconstruction quality for reflective objects in uncontrolled real-world scenarios. It includes three challenging scenes, \textit{i.e.,} ``sedan", ``garden spheres", and ``toycar", each featuring multi-view images of highly reflective surfaces. Ref-NeRF specifically tests an algorithm's robustness to specular reflections.
	(\textbf{iv}) \textbf{DeepVoxels}\cite{Sitzmann_Thies_Heide_NieBner_Wetzstein_Zollhofer_2019}: To evaluate performance in more realistic settings, we select two scenes (a glossy sphere and a coffee cup) from this dataset. Both scenes were captured with handheld cameras and processed via COLMAP~\cite{schoenberger2016sfm, schoenberger2016mvs}, featuring reflective surfaces under non-ideal conditions.

	\noindent\textbf{Evaluation Metrics.}
	For scenes with available ground-truth geometry (\textit{e.g.}, DTU dataset), we employ the Chamfer Distance (CD) as the primary quantitative metric to assess the accuracy of the reconstructed point cloud. Following the official evaluation protocol, CD is defined as:
	\begin{equation}
		CD_{(\boldsymbol{A}, \boldsymbol{B})} = \frac{1}{|\boldsymbol{A}|} \sum_{\boldsymbol{a} \in\boldsymbol{ A}} \min_{\boldsymbol{b} \in \boldsymbol{B}} \| \boldsymbol{a} - \boldsymbol{b} \|_2^2 + \frac{1}{|\boldsymbol{B}|} \sum_{\boldsymbol{b} \in \boldsymbol{B}} \min_{\boldsymbol{a} \in \boldsymbol{A}} \| \boldsymbol{b} -\boldsymbol{a} \|_2^2,
	\end{equation}
	where $\boldsymbol{A}$ and $\boldsymbol{B}$ are the predicted and ground-truth point sets, respectively. The first term measures how close the prediction is to the ground truth (accuracy), and the second term measures coverage of the ground truth by the prediction (completeness). Lower CD values indicate higher geometry fidelity and are reported in millimeters for DTU following the official benchmark.
	For the Shiny Blender dataset, we follow UniSDF~\cite{wang2023unisdf} and Ref-NeuS~\cite{ge2023ref-neus} to report mean angular error (MAE) of normals and accuracy (ACC) of mesh. The ground-truth meshes are double-layered; the inner layer is not visible in multi-view images and cannot be reconstructed, so the completeness term of the Chamfer Distance would be biased by these invisible points. We therefore report only ACC, which corresponds to the accuracy term (first term) in the CD definition above and reflects how well the predicted surface aligns with the visible ground truth (lower is better). We also evaluate surface orientation with the MAE between rendered normals and ground-truth normals over the valid foreground mask. Consistent with Ref-NeuS, ACC is not reported for the ball and teapot scenes.
	For datasets lacking ground-truth geometry (\textit{e.g.}, Ref-NeRF and DeepVoxels), we conduct a qualitative evaluation. We focus on the preservation of fine geometric details, especially for reflective surfaces, such as the accurate reconstruction of depressions, as well as the overall reconstruction plausibility in both reflective and non-reflective areas.
	
	\noindent\textbf{Implementation Details.} 
	\label{Implementation-Details}
	We implement our algorithm in PyTorch. 
	For efficiency, we adopt the approach from \cite{lee2022tracer} to obtain image mask information.
	The training process follows our four-stage training strategy described in Section~\ref{sec:training-strategy}, with the following iteration schedule: we start with geometric initialization for 2,500 iterations. This is followed by the first level of training for 5,000 iterations, then a coarse-to-fine stage for 10,000 iterations divided into 4 levels with 2,500 iterations per level, and finally, the finest level training for 20,000 iterations. 
	In total, the training comprises 37,500 iterations.
	All experiments are conducted on an NVIDIA A100 GPU (40GB), taking approximately 40 minutes for training (excluding pose estimation and data preprocessing time). 
	After training, we extract the final model using the Marching Cubes algorithm~\cite{Lorensen_2020} with a resolution of 1024.
	
	\noindent\textbf{Sampling Strategy.}
	Following the sampling strategy proposed in \cite{Barron_Mildenhall_Verbin_Srinivasan_Hedman_2022}, we employ two rounds of proposal sampling and one NeRF sampling, with sample counts of 256, 96, and 24, respectively. More specifically,
	the final rendering is determined by the 24 sampling points from the NeRF sampling. 
	Notably, to focus on object-level reconstruction, we do not perform sampling along the ray path from its origin until it intersects with the axis-aligned bounding box.

	\begin{figure*}[t]
		\centering
		\includegraphics[width=0.95\linewidth]{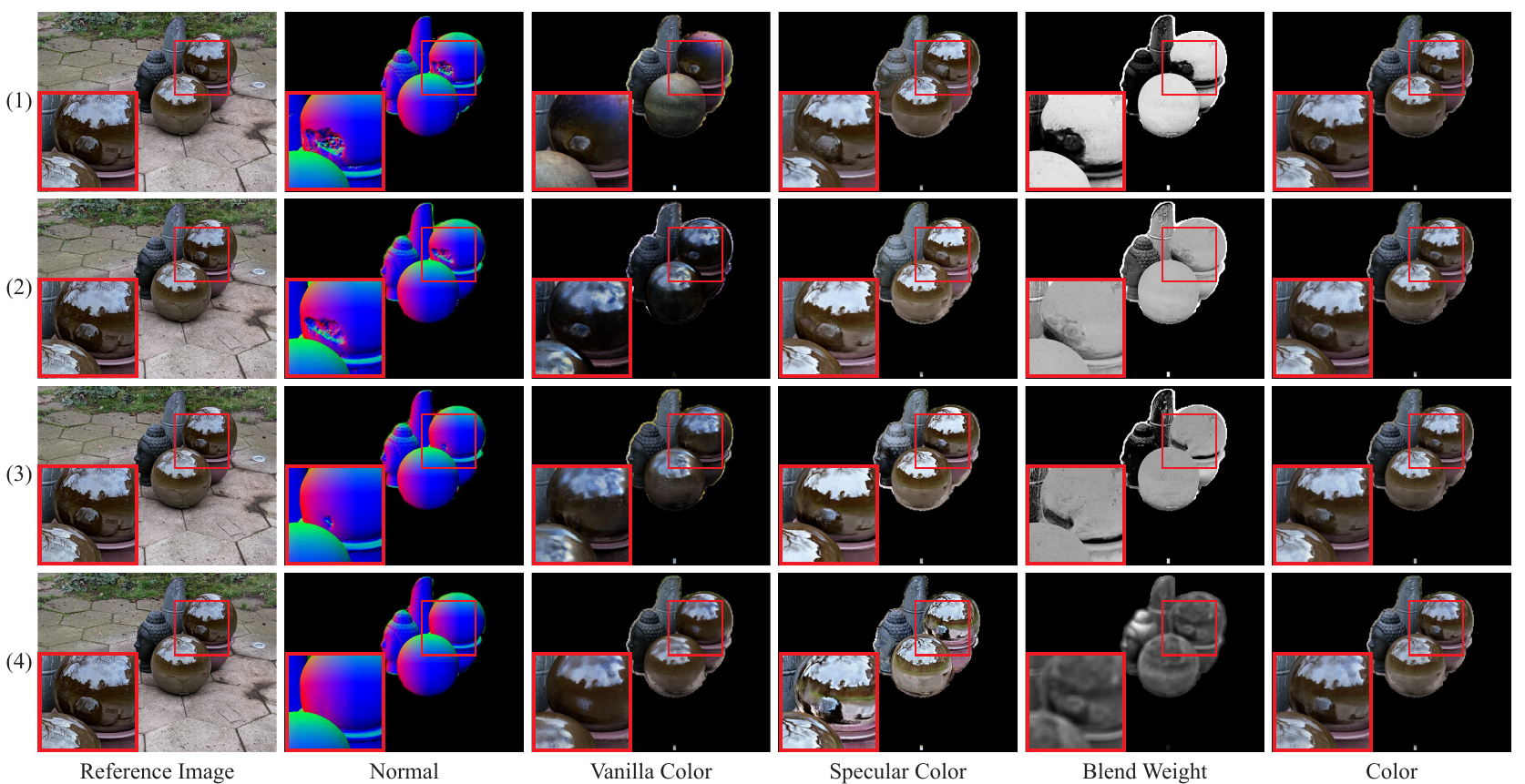}
		\vspace{-0.2cm}
		\caption{
			Qualitative results of the proposed method under different configurations: (1) Baseline; (2) Baseline + RD (reflection direction for vanilla MLP input); (3) Baseline + RD + MS (microfacet sampling); (4) Full model: baseline + RD + MS + DC (detection-based color blending). }
		\label{fig:gardensphere-ablation}
		\vspace{-0.2cm}
	\end{figure*}

	\subsection{Comparison with the State-of-the-Art}

	\noindent\textbf{DTU Dataset.}
	As shown in~Table~\ref{tab:dtu-results}, our proposed method achieves the best performance on most scenes and the best average Chamfer distance among all compared methods, demonstrating strong generalization across scenes with both reflective and non-reflective surfaces. The comparison comprises neural implicit methods (e.g., NeuS~\cite{NEURIPS2021_e41e164f}, UniSDF~\cite{wang2023unisdf}) and recent 3D Gaussian Splatting based methods (SuGaR~\cite{guedon2024sugar}, 2DGS~\cite{huang20242d}, GOF~\cite{yu2024gaussian}, Rade-GS~\cite{zhang2024rade}, GSDF~\cite{yu2024gsdf}). Optimization time in the table ranges from $\sim$0.5\,h to 16\,h per scene: several 3DGS methods require less time (e.g., $\sim$0.5\,h), while some neural implicit methods require over 16\,h. Our method takes $\sim$1\,h per scene and attains the best mean Chamfer distance, offering a favorable trade-off between efficiency and accuracy.
	Fig.~\ref{fig:dtu-results} presents qualitative comparisons, which further illustrate the superiority of our reconstructions. Our method consistently recovers finer geometry details, such as the facial features of the DTU scan118 statue, more accurately than NeuS\cite{NEURIPS2021_e41e164f} (which requires over ten hours of optimization) and significantly surpasses UniSDF\cite{wang2023unisdf}, a method designed for general object reconstruction.
	
	\noindent\textbf{Shiny Blender Dataset.}
	We compare our method with recent rendering-oriented work (Ref-NeRF~\cite{2022refnerf}), inverse rendering (NMF~\cite{mai2023neural}), conventional neural implicit methods (Geo-NeuS~\cite{fu2022geo}, Neuralangelo~\cite{li2023neuralangelo}), and methods designed for reflective surfaces (Ref-NeuS~\cite{ge2023ref-neus}, UniSDF~\cite{wang2023unisdf}). As shown in Table~\ref{tab:shiny_blender}, our method achieves the best mean MAE (4.58$^\circ$) and the best mean ACC (0.61) among compared methods. This demonstrates the advantages of the proposed microfacet-regularized SDF representation for high-precision geometry reconstruction in reflective scenes.
	
	\noindent\textbf{Ref-NeRF and DeepVoxels Datasets.} 
	As shown in Fig.~\ref{fig:ref-nerf-results}. We compare the qualitative reconstruction results of our method against those of Ref-NeuS~\cite{ge2023ref-neus}, UniSDF~\cite{wang2023unisdf} (via InstantAngelo~\cite{hugoycj_instant_angelo}), and AniSDF~\cite{gao2024anisdf}. These datasets pose significant challenges due to uncontrolled lighting, real-world reflections, and specular highlights. Our method demonstrates robust performance under these conditions, consistently preserving complex geometric structures and accurately reconstructing specular regions.
	Specifically, Ref-NeuS struggles on real-world reflective surfaces, such as in the ``toycar" scene where it incorrectly reconstructs depressed regions as flat surfaces. AniSDF and UniSDF show improved performance over Ref-NeuS in handling reflective surfaces but still fall short in preserving fine geometric details. 
	AniSDF particularly suffers from texture replication artifacts, erroneously interpreting printed patterns as geometric features and exhibiting a ``wrapped" appearance of the target object, as seen in the first two scenes (from top to bottom) in Fig.~\ref{fig:ref-nerf-results}. In contrast, our approach maintains geometric details while avoiding texture replication artifacts, achieving more accurate and physically plausible reconstructions.

	\begin{figure*}[t]
		\centering	\includegraphics[width=0.95\linewidth]{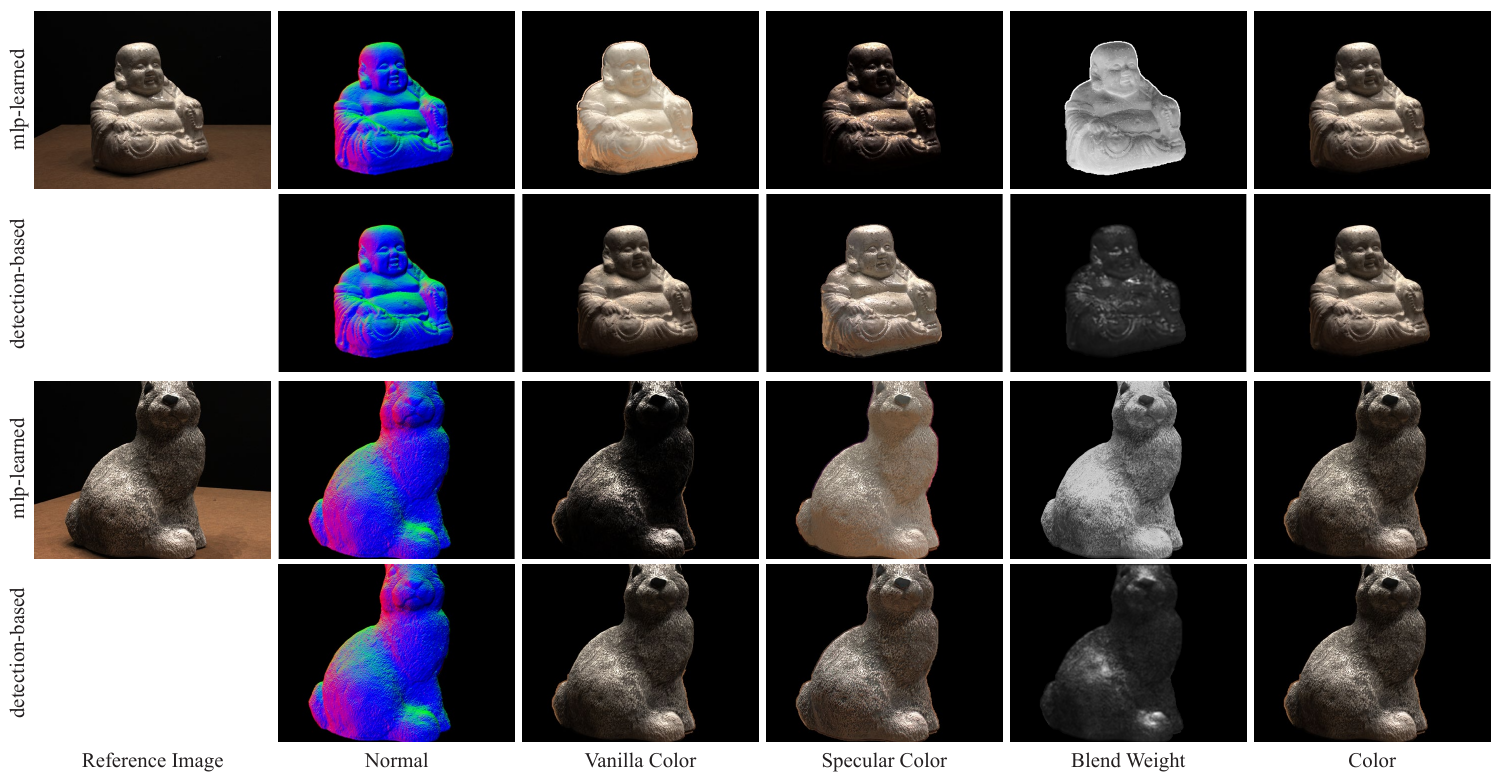}
		\vspace{-0.2cm}
		\caption{Qualitative comparison between the MLP-learned and the proposed detection-guided color blending methods on the DTU dataset.}
		\label{fig:ablation-color_blend}
		\vspace{-0.2cm}
	\end{figure*}
	
	\begin{figure*}[t]
		\centering
		\includegraphics[width=0.95\linewidth]{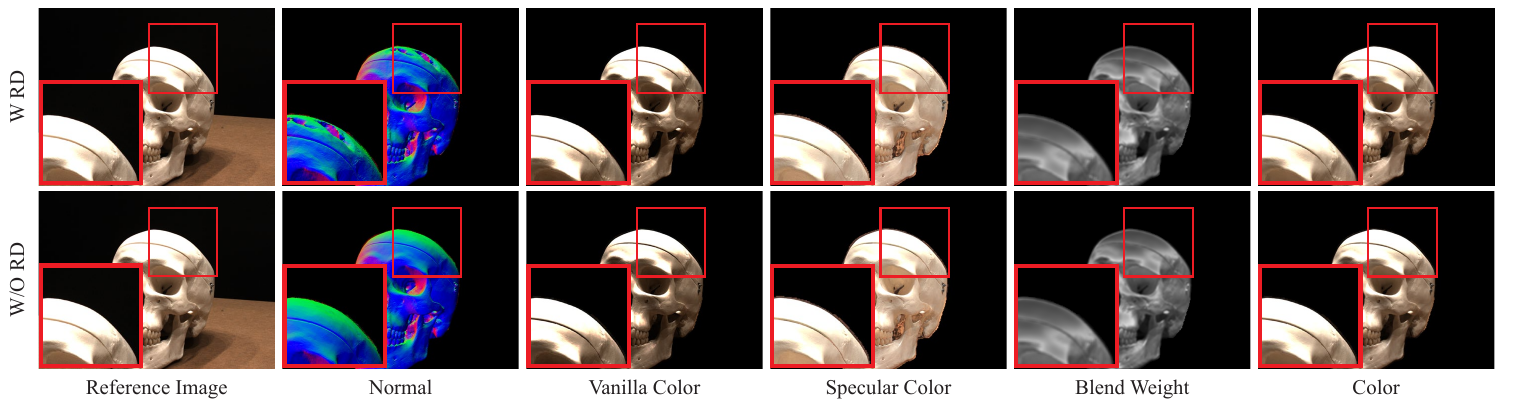}
		\vspace{-0.2cm}
		\caption{Qualitative comparison between using view direction and reflective direction as input to the vanilla MLP of the color network.}
		\label{fig:ablation_view_dir_vs_ref_dir}
		\vspace{-0.2cm}
	\end{figure*}

	\begin{figure*}[t]
		\centering
		\includegraphics[width=1.0\linewidth]{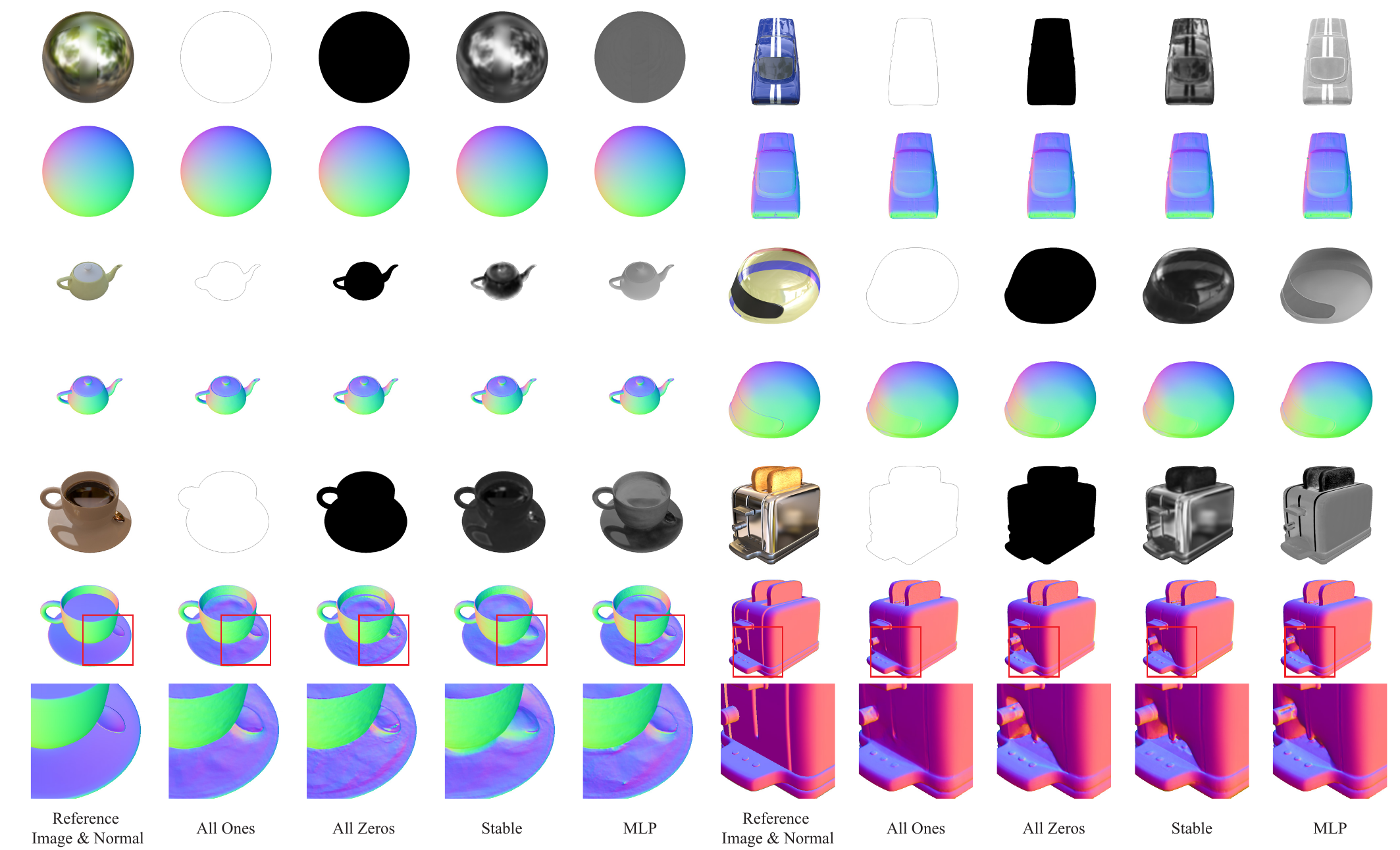}
		\caption{Ablation of specular blending weights on the Shiny Blender dataset. Columns: ``Reference Image and Normal";``All Ones" (weights 1, specular branch only); ``All Zeros" (weights 0, diffuse branch only); ``Stable" (StableDelight detector); ``MLP" (learned weights as in UniSDF). For each object, the first row shows the blending weight map, the second row shows the reconstructed normal map.}
		\label{fig:dc_ablation}
		\vspace{-0.3cm}
	\end{figure*}

	\begin{figure}[t]
		\centering
		\includegraphics[width=1.0\linewidth]{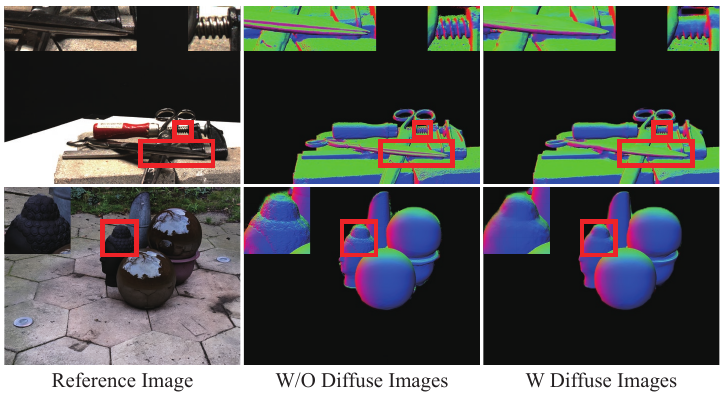}
		\vspace{-0.5cm}
		\caption{Ablation result on diffuse images, showcasing reconstruction of reflective surfaces.}
		\label{fig:ablation-diffuse-images}
		\vspace{-0.5cm}
	\end{figure}

	\subsection{Ablation Studies}
	\label{sec:ablation-studies}
	
	\noindent\textbf{Component-wise Analysis of Color Modeling.}
	Compared to the color network configuration in UniSDF~\cite{wang2023unisdf}, our approach introduces three key modifications: (\textbf{i}) Incorporating the \textbf{reflection direction} (RD) as input to the Vanilla MLP; (\textbf{ii}) Introducing \textbf{microfacet sampling} (MS) during training, and (\textbf{iii}) Utilizing a \textbf{detection-based specular score} (DC) as the color blending weight between the vanilla and specular branches.
	To investigate their individual and combined effects of these components, we start from a baseline configuration that excludes all three components and gradually add each one. 
	
	Qualitative results in Fig.~\ref{fig:gardensphere-ablation} illustrate the individual contributions of each component in our method. 
	The baseline model, without RD, MS, and DC, fails to reconstruct high-frequency, nearby reflective surfaces accurately, resulting in distorted geometry, such as the region highlighted in red in Fig.~\ref{fig:gardensphere-ablation} , which reflects the nearby sphere like a mirror.
	Introducing RD improves surface continuity by incorporating view-dependent cues. Adding MS further enhances geometric fidelity, particularly in reflective regions. Finally, the full model incorporating RD, MS, and DC achieves a complete and accurate reconstruction of the spherical object, demonstrating the effectiveness of the proposed design.
	
	Quantitative results on the DTU dataset are presented in Table~\ref{tab:ablation-combined}, showing the 
	contributions of each component via Chamfer Distance (CD, lower is better). 
	Combining all modifications (MS, RD, DC) yields an average CD of 0.607. Removing MS, RD, or DC individually degrades performance by 7.08\% (CD 0.65), 5.44\% (CD 0.64), or 1.32\% (CD 0.615), respectively.
	While the DC component exhibits relatively smaller performance gains on the DTU dataset when combined with the MS and RD components, compared to the baseline that learns color blending weights automatically via an MLP, it remains valuable from a physical intuition perspective.
	Besides, its removal significantly worsens the CD on metallic objects like DTU 110 by 13.64\% (from 0.66 to 0.75).
	Specifically, explicitly introducing detection-based blending weights provides more interpretable and physically grounded guidance for the color branches. 
	In contrast, MLP-learned weights may overfit to dataset-specific cues, leading to less physically meaningful blending behaviors. 
	As illustrated in the two DTU scenes shown in Fig.~\ref{fig:ablation-color_blend}, the MLP-learned approach erroneously predicts predominantly specular reflections for surfaces with diffuse characteristics, deviating from human visual perception.
	Fig.~\ref{fig:ablation_view_dir_vs_ref_dir} provides a qualitative visualization for the DTU scene 65 results in Table~\ref{tab:ablation-combined}, revealing that while RD has minimal impact on color appearance, it significantly improves geometric reconstruction, particularly in the bright white region on the skull's crown.

	\begin{table*}[t]
		\caption{Impact of the microfacet normal consistency loss (MNCL) on the DTU dataset (Chamfer Distance (mm) $\downarrow$). We ablate MNCL while keeping MS, RD, and DC enabled. \colorbox{red!25}{Red} denotes the best results.}
		\label{tab:ablation-mncl}
		\vspace{-0.2cm}
		\centering
		\small
		\begin{tabularx}{\textwidth}{l *{16}{>{\centering\arraybackslash}X}}
			\toprule
			MS & 24 & 37 & 40 & 55 & 63 & 65 & 69 & 83 & 97 & 105 & 106 & 110 & 114 & 118 & 122 & Mean \\
			\midrule
			$\times$ & 0.95 & \cellcolor{red!25}0.62 & 0.32 & 0.51 & 0.92 & \cellcolor{red!25}0.58 & \cellcolor{red!25}0.52 & 1.37 & 1.01 & \cellcolor{red!25}0.51 & 0.50 & 0.69 & 0.35 & 0.43 & \cellcolor{red!25}0.39 & 0.65\\
			\checkmark (wo MNCL) & 0.82 & 0.67 & 0.38 & 0.50 & \cellcolor{red!25}0.83 & 0.63 & 0.60 & 1.36 & \cellcolor{red!25}0.96 & 0.53 & 0.50 & 0.67 & 0.38 & 0.44 & 0.40 & 0.65 \\
			\checkmark (w MNCL) & \cellcolor{red!25}0.53 & 0.64 & \cellcolor{red!25}0.31 & \cellcolor{red!25}0.44 & 0.88 & 0.62 & 0.56 & \cellcolor{red!25}1.34 & 1.00 & 0.52 & \cellcolor{red!25}0.48 & \cellcolor{red!25}0.66 & \cellcolor{red!25}0.33 & \cellcolor{red!25}0.39 & 0.40 & \cellcolor{red!25}0.607 \\
			\bottomrule
		\end{tabularx}
		\vspace{-0.5cm}
	\end{table*}

	\begin{table*}[t]
		\caption{Comparison between Dense Initialization and Final Reconstruction Results on DTU. All metrics are reported in Chamfer Distance (mm) $\downarrow$ (lower is better). \colorbox{red!25}{Red} denotes the best results.}
		\label{tab:Comparison-GI}
		\vspace{-0.2cm}
		\centering
		\begin{tabularx}{\textwidth}{l *{16}{>{\centering\arraybackslash}X}}
			\toprule
			\makecell{Method} & 24 & 37 & 40 & 55 & 63 & 65 & 69 & 83 & 97 & 105 & 106 & 110 & 114 & 118 & 122 & Mean \\
			\midrule
			OpenMVS\cite{openmvs_github} & \cellcolor{red!25}0.53 & 0.87 & 0.34 & \cellcolor{red!25}0.38 & 1.12 & 1.01 & 0.65 & 1.48 & 1.10 & 0.72 & 0.51 & 0.93 & 0.35 & 0.48 & 0.51 & 0.73 \\
			Ours &  \cellcolor{red!25}0.53 & \cellcolor{red!25}0.64 & \cellcolor{red!25}0.31 & 0.44 & \cellcolor{red!25}0.88 & \cellcolor{red!25}0.62 & \cellcolor{red!25}0.56 & \cellcolor{red!25}1.34 & \cellcolor{red!25}1.00 & \cellcolor{red!25}0.52 & \cellcolor{red!25}0.48 & \cellcolor{red!25}0.66 & \cellcolor{red!25}0.33 & \cellcolor{red!25}0.39 & \cellcolor{red!25}0.40 & \cellcolor{red!25}0.61 \\
			\bottomrule
		\end{tabularx}
		\vspace{-0.5cm}
	\end{table*}
	
	\begin{figure*}[t]
		\centering
		\includegraphics[width=0.95\linewidth]{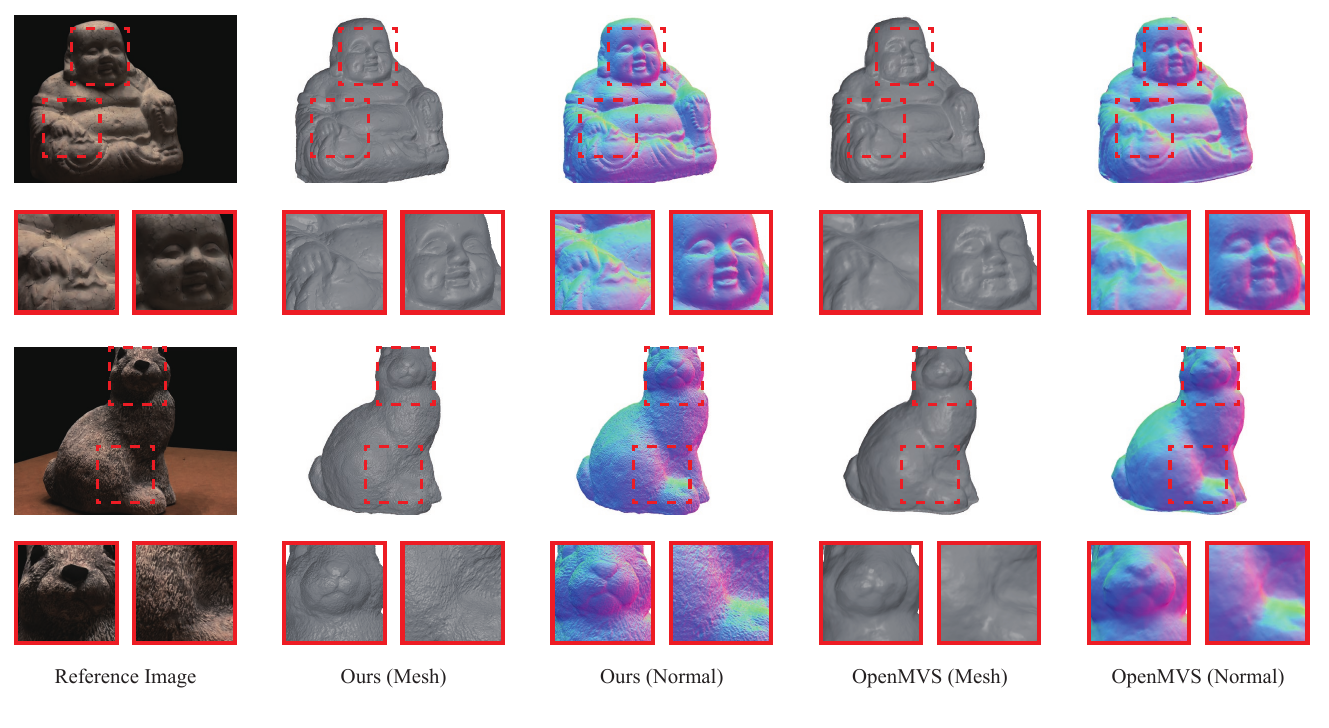}
		\vspace{-0.5cm}
		\caption{Qualitative comparison between OpenMVS\cite{openmvs_github} and ours on the scan 55 and scan 114 from the DTU dataset.}
		\label{fig:compar-openmvs-final-result}
		\vspace{-0.2cm}
	\end{figure*}

	\begin{table*}[t]
		\caption{Ablation Study on Geometry Initialization (GI), Microfacet Sampling Scale (MSS) and Diffuse Images (DI). All metrics are reported in Chamfer Distance (mm) $\downarrow$ (lower is better). \colorbox{red!25}{Red} and \colorbox{orange!25}{orange} denote the best and second-best per column. ``cur+4'' denotes a scale that is 4 levels finer than the current level, and ``cur+8'' denotes a scale that is 8 levels finer than the current level (i.e., the finest granularity level in our setting).}
		\label{tab:ablation-study-GI-MN-scale}
		\centering
		\begin{tabularx}{\textwidth}{l l l *{17}{>{\centering\arraybackslash}X}}
			\toprule
			& \makecell{GI} & \makecell{MSS} & \makecell{DI} & 24 & 37 & 40 & 55 & 63 & 65 & 69 & 83 & 97 & 105 & 106 & 110 & 114 & 118 & 122 & Mean \\
			\midrule
			(a) & Sparse & cur+8 & $\times$  &  0.80 & 0.69 & 0.64 & 1.13 & \cellcolor{red!25}0.55 & 0.64 & 0.61 & \cellcolor{red!25}0.50 & 1.15 & 0.59 & 0.59 & 1.34 & 0.83 & 0.59 & 1.23 & 0.79 \\
			(b) & Sparse & cur+4 & \checkmark & 1.26 & 0.71 & 0.89 & 0.51 & 1.37 & 0.99 & 0.65 & 1.23 & 1.25 & \cellcolor{red!25}0.50 & 0.56 & 0.97 & 0.43 & 0.48 & 0.41 & 0.81 \\
			(c) & Sparse & $\times$  & $\times$ &  0.60 & 0.73 & 0.52 & 0.51 & 1.23 & 0.96 & \cellcolor{orange!25}0.53 & 0.92 & 1.17 & \cellcolor{orange!25}0.51 & 0.54 & 0.70 & 0.36 & \cellcolor{red!25}0.39 & \cellcolor{red!25}0.37 & 0.67 \\
			(d) & Sparse & cur+4 & $\times$ & 0.67 & 0.64 & 0.50 & 0.50 & 1.19 & 0.69 & 0.58 & \cellcolor{orange!25}0.87 & 1.23 & \cellcolor{red!25}0.50 & \cellcolor{red!25}0.48 & \cellcolor{orange!25}0.69 & \cellcolor{orange!25}0.35 & 0.45 & \cellcolor{orange!25}0.38 & 0.65 \\
			(e) & Dense & cur+8 & $\times$ &  \cellcolor{red!25}0.48 & \cellcolor{red!25}0.61 & 0.35 & \cellcolor{orange!25}0.46 & \cellcolor{orange!25}0.88 & 0.64 & 0.55 & 1.34 & 1.03 & 0.57 & 0.52 & \cellcolor{orange!25}0.69 & 0.37 & \cellcolor{red!25}0.39 & 0.40 & \cellcolor{orange!25}0.62 \\
			(f) & Dense & cur+4 & \checkmark & 0.65 & 0.64 & 0.41 & 0.50 & 1.02 & 0.95 & 0.65 & 1.53 & 1.10 & 0.52 & 0.57 & 0.93 & 0.41 & 0.50 & 0.44 & 0.72 \\
			(g) & Dense & $\times$ & $\times$  & 0.95 & \cellcolor{orange!25}0.62 & \cellcolor{orange!25}0.32 & 0.51 & 0.92 & \cellcolor{red!25}0.58 & \cellcolor{red!25}0.52 & 1.37 & \cellcolor{orange!25}1.01 & \cellcolor{orange!25}0.51 & \cellcolor{orange!25}0.50 & \cellcolor{orange!25}0.69 & \cellcolor{orange!25}0.35 & \cellcolor{orange!25}0.43 & 0.39 & 0.64 \\
			\textbf{Ours} & Dense & cur+4 & $\times$ & \cellcolor{orange!25}0.53 & 0.64 & \cellcolor{red!25}0.31 & \cellcolor{red!25}0.44 & \cellcolor{orange!25}0.88 & \cellcolor{orange!25}0.62 & 0.56 & 1.34 & \cellcolor{red!25}1.00 & 0.52 & \cellcolor{red!25}0.48 & \cellcolor{red!25}0.66 & \cellcolor{red!25}0.33 & \cellcolor{red!25}0.39 & 0.40 & \cellcolor{red!25}0.61 \\
			\bottomrule
		\end{tabularx}
		\vspace{-0.5cm}
	\end{table*}
	\noindent\textbf{Effect of Specular Blending Weights.} To further quantify the impact of specular blending weights on reconstruction performance, we conduct an ablation study on the Shiny Blender dataset and DTU dataset by varying how these weights are defined. 
	We compare four specular blending weight configurations: ``Stable'' (StableDelight detection scores directly), ``All Ones'' (all pixels treated as specular), ``All Zeros'' (all pixels treated as diffuse), and ``MLP'' (blending weights learned via an auxiliary MLP as in UniSDF~\cite{wang2023unisdf}). 
	
	As shown in Table~\ref{tab:shiny_blender_dc}, on Shiny Blender, which predominantly contains reflective materials, ``All Ones'' achieves the best performance (mean MAE 4.58$^\circ$, mean ACC 0.61), consistent with the dataset statistics. ``Stable'' performs relatively worse (mean MAE 5.37$^\circ$). This is due to the large domain gap between the pretrained detector and this synthetic reflective domain.
	However, even under this unsuitable prior, the model remains stable and competitive with other state-of-the-art methods (MAE ranks the top three, ACC remains the best as shown in Table~\ref{tab:shiny_blender}), indicating that the framework does not collapse under imperfect specular guidance. Qualitatively (Fig.~\ref{fig:dc_ablation}), the advantage of stronger specular emphasis (``All Ones'') is particularly evident in near-field reflection regions (e.g., Toaster as shown in the 7th row 7th column of Fig.~\ref{fig:dc_ablation}), whereas ``Stable'' underestimates reflective confidence in certain scenes (e.g., Coffee, as shown in 7th row 4th column of Fig.~\ref{fig:dc_ablation}), leading to less accurate normal reconstruction. The ``MLP'' configuration yields balanced performance but does not fully match the reflective-only prior in highly specular regions. As shown in Table~\ref{tab:dtu_dc}, on DTU, which contains mixed reflective and non-reflective materials, ``Stable'' achieves the best mean Chamfer Distance (0.607), outperforming fixed priors (``All Ones'': 0.65; ``All Zeros'': 0.63) and learned blending (``MLP'': 0.615). This indicates that detector-derived blending weights are beneficial when reflective and diffuse regions coexist.
	
	These results show that performance is affected by the suitability of the specular prior to the scene statistics rather than a hard dependency on detector accuracy. The detector serves as a soft, replaceable guidance signal, and the framework remains robust under extreme blending configurations.
	
	\noindent\textbf{Effect of Microfacet Normal Consistency Loss.}
	The microfacet normal consistency loss (MNCL) enforces the alignment between microfacet normals and local surface normals in specular regions (as detailed in Section III-B3 and Eq. 7.), reflecting the core principle of microfacet theory.
	To isolate the impact of this loss term, we ablate it while keeping the full configuration (MS, RD, DC as shown in Table~\ref{tab:ablation-combined}). As shown in Table~\ref{tab:ablation-mncl}, removing MNCL degrades performance by 7.08\% (CD 0.65). 
	This demonstrates that the consistency loss plays a crucial role in maintaining physically plausible surface orientations, which directly benefits the reconstruction quality.
	
	\noindent\textbf{On the Use of Diffuse \textit{v.s.} Original Images in the Vanilla branch.}
	To evaluate the suitability of using diffuse images as input for the vanilla branch of our dual-branch color network, we conducted a comparative experiment with two types of input: diffuse images and original images. 
	As shown in Fig.~\ref{fig:ablation-diffuse-images}, although directly employing diffuse images for reconstruction may intuitively seem appropriate for modeling diffuse reflectance, our results reveal that they significantly limit reconstruction quality. 
	Specifically, using diffuse images leads to a substantial loss of detail, which in turn compromises the geometric fidelity of the reconstructed models. This results in degraded reconstructions, particularly in fine structural features, such as the blade edges of the scissors and the helical patterns of the screwdriver in DTU scan 37, as well as the intricate details of the buddha statue in the ``gardensphere" scene. In these cases, the loss of surface detail significantly impacts reconstruction quality. Quantitatively, as shown in Table~\ref{tab:ablation-study-GI-MN-scale}, the CD metric drops from 0.81 to 0.65 between (b) and (d), and from 0.72 to 0.61 between (f) and Ours, further confirming the adverse effect of using diffuse images directly.

\begin{figure}[t]
		\centering
		\includegraphics[width=1.0 \linewidth]{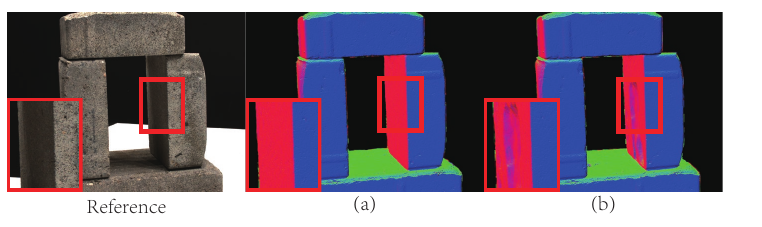}
		\vspace{-0.8cm}
		\caption{Comparison of reconstruction results using different initialization strategies: (a) dense point cloud initialization from OpenMVS, and (b) sparse point cloud initialization from COLMAP.}
		\label{fig:ablation-dense-and-sparse-initialization-ablation}
		\vspace{-0.3cm}
	\end{figure}

	\noindent\textbf{Geometry Initialization Strategies.}
	\label{ablation-gi}
	To evaluate the impact of initialization points on reconstruction results, we consider using initialization points from different sources. 
	Specifically, we conduct experiments on the DTU dataset, comparing two initialization strategies: using sparse initialization points and using dense initialization points. 
	The sparse initialization points are byproducts obtained during COLMAP's camera pose estimation, requiring no additional processing. However, the resulting point cloud is unevenly distributed and of limited quality.
	In contrast, dense initialization points are generated using downsampled image sets by OpenMVS~\cite{openmvs_github}, which, while requiring more computational overhead compared to COLMAP, still represents a relatively small portion of the overall reconstruction process (approximately a few minutes). 
	We utilize traditional 3D reconstruction results as the starting point for first-stage geometric initialization. 
	To minimize the potential influence of initialization results on subsequent performance, we release the constraints of dense input points during the following three stages. 
	The quantitative comparison results with OpenMVS are shown in Table~\ref{tab:Comparison-GI}.
	Our method consistently outperforms OpenMVS across all scenes except for DTU scan 55, where OpenMVS reports a slightly better metric score.
	However, as illustrated in Fig.~\ref{fig:compar-openmvs-final-result}, our method produces visibly superior results on this scene, with significantly better preservation of fine details and surface quality.
	As shown in Fig.~\ref{fig:ablation-dense-and-sparse-initialization-ablation}, 
	the reconstruction results using dense initialization appear significantly smoother compared to those using sparse initialization, especially in regions with darker colors.
	This conclusion is further validated in Table~\ref{tab:ablation-study-GI-MN-scale}; for example, the CD metric improved from 0.67 (d) to 0.61 (Ours).

	\noindent\textbf{Effect of Microfacet Normal Scale.}
	We evaluate the performance of microfacet normals at different scales under both sparse and dense input conditions, as shown in Table~\ref{tab:ablation-study-GI-MN-scale}. In our setting, the initial level is 8 and the finest level is 16. We compare two scale choices. ``cur+4'' denotes a scale 4 levels finer than the current level, and ``cur+8'' denotes 8 levels finer, i.e.\ the finest granularity, corresponding to directly sampling microfacet normals at level 16. Under dense input, switching from the finest scale to ``cur+4'' improves CD from 0.62 in row (e) to 0.61 in Ours. Under sparse input, the same switch from ``cur+8'' to ``cur+4'' improves CD from 0.79 in row (a) to 0.65 in row (d), a larger gain than in the dense case. 
	In both conditions, ``cur+4'' outperforms the finest scale ``cur+8''.
	This suggests that using the finest grid scale is not necessarily optimal, which is particularly evident when the geometry initialization is sparse, where overly fine scales may amplify noise or lead to unstable normal estimation.
	
	\section{Conclusion}
	In this work, we presented MicroSDF, a novel framework for 3D reconstruction of scenes with both reflective and non-reflective surfaces.
	Grounded in physical principles, MicroSDF integrates microfacet theory to guide both geometric and color modeling. Specifically, we introduce a microfacet consistency loss to enforce physically meaningful surface geometry for reflective regions, while leveraging microfacet normals to facilitate specular color learning. This is further enhanced by a detection-guided color blending strategy, which provides external supervision for physically plausible color modeling.
	Together, these components enable our framework to effectively handle both reflective and non-reflective surfaces in a unified manner.
	
	Extensive experiments across diverse benchmark
	datasets validate the effectiveness and robustness of our framework. On the DTU dataset, which mixes reflective and non-reflective surfaces, MicroSDF achieves state-of-the-art performance with an average Chamfer Distance of 0.61, outperforming existing methods. On the synthetic reflective Shiny Blender benchmark, our method also attains state-of-the-art quantitative results, demonstrating its strong capability in handling highly specular scenes. For real-world reflective datasets such as Ref-NeRF and DeepVoxels, we provide qualitative evaluations that show high-fidelity reconstruction of reflective regions and improved surface consistency. These results indicate that MicroSDF generalizes well across both controlled synthetic environments and real-world scenarios. Moreover, the proposed integration of microfacet-driven priors offers a principled way to incorporate physically grounded reflectance modeling into neural reconstruction frameworks. This idea is potentially compatible with recent 3D Gaussian Splatting approaches for improved reconstruction of reflective and specular surfaces, which we consider a promising direction for future work.
	
	\vspace{-0.3cm}
	\bibliographystyle{IEEEtran}
	\bibliography{references}
\end{document}